\documentclass[11pt]{article}

\usepackage[preprint]{acl}

\usepackage{times}
\usepackage{latexsym}

\usepackage[T1]{fontenc}

\usepackage[utf8]{inputenc}

\usepackage{microtype}

\usepackage{inconsolata}

\usepackage{graphicx}
\usepackage{booktabs}       %
\usepackage{amsfonts}       %
\usepackage{amsmath}        %
\usepackage{nicefrac}       %
\usepackage{xcolor}         %
\usepackage{amsmath}
\usepackage{graphicx}
\usepackage{subcaption} 
\usepackage{algorithm}
\usepackage{algorithmic}
\usepackage{multirow}
\usepackage{amsfonts}       %
\usepackage{amsmath}        %
\usepackage{nicefrac}       %
\usepackage{microtype}      %
\usepackage{xcolor}         %
\usepackage{amsmath}
\usepackage{graphicx}
\usepackage{subcaption} 
\usepackage{algorithm}
\usepackage{algorithmic}
\usepackage{multirow}
\usepackage{xcolor}
\usepackage{pifont}
\usepackage{arydshln}

\newcommand{\cmark}{\ding{51}} 
\newcommand{\xmark}{\ding{55}} 
\newcommand{\method}{ScenGround}

\title{Beyond Referring Expressions: Scenario Comprehension Visual Grounding}

\author{
Ruozhen He\textsuperscript{1},
Nisarg A. Shah\textsuperscript{2},
Qihua Dong\textsuperscript{3},
Zilin Xiao\textsuperscript{1},
Jaywon Koo\textsuperscript{1},
Vicente Ordonez\textsuperscript{1} \\
\textsuperscript{1}Rice University \quad
\textsuperscript{2}Johns Hopkins University \quad
\textsuperscript{3}Northeastern University \\
\texttt{\{catherine.he,zilin,jk125,vicenteor\}@rice.edu} \\
\texttt{snisarg812@gmail.com} \quad
\texttt{dongqh078@gmail.com}
}

\begin{document}
\maketitle

\begin{abstract}
Existing visual grounding benchmarks primarily evaluate alignment between image regions and literal referring expressions, where models can often succeed by matching a prominent named category. We explore a complementary and more challenging setting of scenario-based visual grounding, where the target must be inferred from roles, intentions, and relational context rather than explicit naming. We introduce Referring Scenario Comprehension (RSC), a benchmark designed for this setting. 
The queries in this benchmark are paragraph-length texts that describe object roles, user goals, and contextual cues that include deliberate references to distractor objects that often require deep understanding to resolve.
Each instance is annotated with interpretable difficulty tags for uniqueness, clutter, size, overlap, and position which expose distinct failure modes and support fine-grained analysis. 
RSC contains approximately 31k training examples, 4k in-domain test examples, and a 3k out-of-distribution split with unseen object categories. 
We further propose \method, a curriculum reasoning method serving as a reference point for this setting, combining supervised warm-starting with difficulty-aware reinforcement learning.
Experiments show that scenario-based queries expose systematic failures in current models that standard benchmarks do not reveal, and that curriculum training improves performance on challenging slices and transfers to standard benchmarks. 
\end{abstract}

\begin{figure*}[t]
    \centering
    \includegraphics[width=\linewidth]{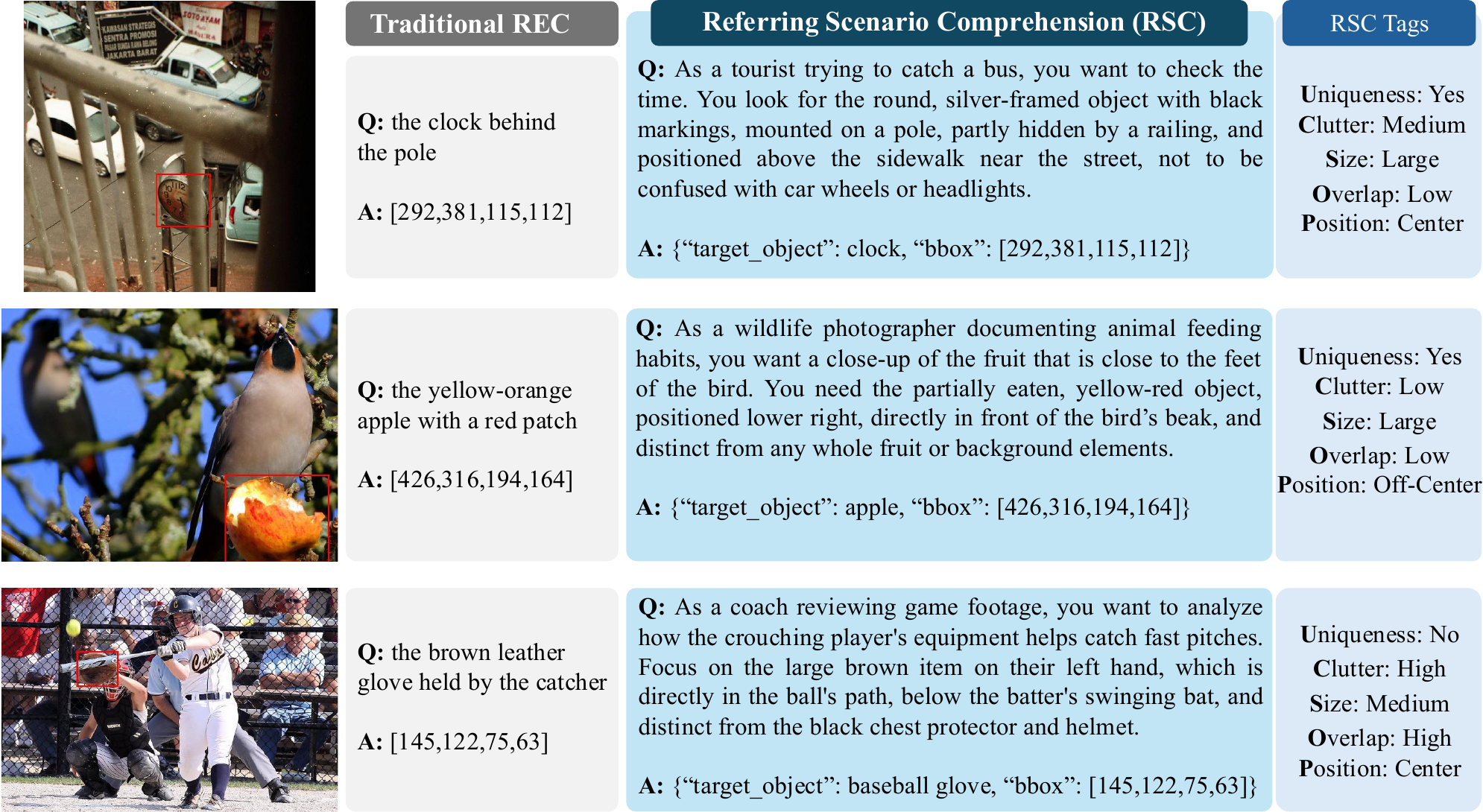}
\caption{\textbf{Referring Scenario Comprehension (RSC) 
vs.\ traditional referring expression comprehension (REC).}
Each row shows the same target object under both paradigms.
Traditional REC queries often name the target category directly, allowing success via lexical matching.
RSC instead pairs each image with a lengthy scenario-based query specifying a user role, goal, and multiple disambiguating cues, including explicit contrasts against competing objects, and requires output identifying both the target object and its bounding box.
The RSC difficulty tags (U/C/S/O/P: Uniqueness, Clutter, Size, Overlap, Position) characterize each instance, enabling fine-grained training and evaluation.
}
\vspace{-0.2in}
    \label{fig:teaser}
\end{figure*}

\begin{table*}[t]
\fontsize{7.5}{9}\selectfont
\centering
\renewcommand{\arraystretch}{1.2}  
\setlength{\tabcolsep}{2pt}
\caption{
\textbf{Comparison of referring expression grounding benchmarks.}
We compare existing datasets with RSC across dataset scale, evaluation splits, referring/query style, reasoning supervision, difficulty annotations, and competing-object mentions.
Most prior datasets rely on short literal phrases and lack explicit reasoning traces or OOD evaluation.
RSC introduces scenario-based queries written as natural language paragraphs and provides per-instance reasoning trace annotations. It further labels five interpretable difficulty factors (U/C/S/O/P) and includes both in-distribution (ID) and out-of-distribution (OOD) category test sets to evaluate reasoning and generalization.
}
\label{tab:benchmark_comparison}
\begin{tabular}{l c c c l l c c c c}
\toprule
\textbf{Benchmark} &
\textbf{Train} &
\textbf{\shortstack{ID \\ Test}} &
\textbf{\shortstack{OOD \\ Test}} &
\textbf{\shortstack{Referring \\ Style}} &
\textbf{\shortstack{Query \\ Style}} &
\textbf{\shortstack{Avg.\\ $|q|$}} &
\textbf{\shortstack{Reasoning \\ Traces}} &
\textbf{\shortstack{Difficulty \\Tags}} &
\textbf{\shortstack{Competing \\ Mentions}} \\
\midrule
RefCOCO~\cite{kazemzadeh2014referitgame}
  & 120,624 & 10,752 & \xmark 
  & Literal
  & Phrase & 3.5 &
  \xmark &
  \xmark & \xmark \\
RefCOCO+~\cite{kazemzadeh2014referitgame}
  & 120,191 & 10,615 & \xmark 
  & Literal
  & Phrase & 3.5 &
  \xmark &
  \xmark & \xmark \\
RefCOCOg~\cite{mao2016generation}
  & 80,512 & 9,602 & \xmark 
  & Descriptive literal
  & Phrase & 8.2 &
  \xmark &
  \xmark & \cmark \\
Cops-Ref~\cite{Chen_2020_CVPR}
  & 119,603 & 12,586 & \xmark 
  & Compositional literal 
  & Template & 14.4 
  & \xmark
  & \xmark & \cmark \\
Ref-Reasoning~\cite{yang2020graph}
  & 721,164 & 34,609 & \xmark 
  & Literal
  & Template & 8.5 
  & Graph
  & Graph layout & \cmark \\
SK-VG~\cite{chen2023advancing}
  & 23,403 & 6,597  & \xmark 
  & Scene knowledge Q\&A
  & Q\&A &  5.3 %
  & \xmark
  & E/M/H & \cmark \\
FineCops-Ref~\cite{liu2024finecops}
  & 163,792 & 9,605 & \xmark 
  & Compositional literal
  & Phrase & 12.2 
  & \xmark
  & 1/2/3 & \cmark \\
EgoIntention~\cite{sun2025visual}
  & 15,667 & 9,892 & \xmark 
  & Egocentric intention query
  & Sentence & 18.8 
  & \xmark
  & \xmark & \xmark \\
\midrule
\textbf{RSC (Ours)}
  & \textbf{31,342} & \textbf{4,038} & \textbf{3,247}
  & \textbf{Scenario-based query}
  & Paragraph & \textbf{52.7} & Paragraph & U/C/S/O/P & \cmark \\
\bottomrule
\end{tabular}%
\end{table*}

\section{Introduction}

Visual grounding, the task of associating an image region with a natural language description, is a core capability for embodied assistants and multimodal models~\cite{geng2025roboverse, shen2021igibson}. 
Despite rapid progress in building capable vision-and-language models (VLMs), strong performance on standard referring comprehension (REC) often %
relies on 
explicit lexical cues, particularly category references and salient attributes that can be matched to candidate regions without deep understanding~\cite{yu2016refcoco, mao2016generation}. 
This protocol creates a systematic blind spot as models optimized for literal phrase matching may fail when users express a need for an object by describing situations instead of making explicit object references.

Real referring behavior is not necessarily concise and direct. Natural language allows users to describe a target through a situational need, an object role, or a user goal. For example, a user attempting to find what time it is may prompt a model on \emph{``checking the time''} instead of explicitly asking for \emph{``the clock''}. Such scenario-based queries  %
might contain multiple disambiguating cues, demanding reasoning over intent and visual context, not just category lookup. 
Despite being a natural and valid form of reference, this type of queries is absent from existing grounding benchmarks, leaving a gap in how we evaluate and develop grounding models. 
As illustrated in Figure~\ref{fig:teaser}, the same target object are described in completely different ways under traditional REC and scenario-based grounding: the former can succeed via lexical matching, while the latter requires integrating relational, spatial, and intentional cues while actively ignoring explicitly mentioned distractors.

We introduce \textbf{Referring Scenario Comprehension (RSC)}, a benchmark designed to study this under-examined setting. RSC replaces referring phrases with scenario-based queries that describe a user role, goal, and at least three disambiguating cues, and deliberately mentions competing objects to require deep understanding.
Each instance is annotated with reasoning traces and five interpretable difficulty tags (\textbf{U}niqueness, \textbf{C}lutter, \textbf{S}ize, \textbf{O}verlap, and \textbf{P}osition), which expose distinct failure modes and support fine-grained curriculum design and evaluation. 
RSC contains 31k training examples, 4k in-domain test examples, and a 3k out-of-distribution split with unseen object categories, enabling evaluation of both in-domain disambiguation and cross-category generalization.

Table~\ref{tab:benchmark_comparison} situates RSC among existing visual grounding benchmarks. Most prior resources contain short literal phrases, are annotated via crowdsourced expression games or scene-graph templates, and provide no out-of-distribution (OOD) evaluation split. Two recent works move toward richer queries: SK-VG~\cite{chen2023advancing} requires reasoning over external knowledge paragraphs, and EgoIntention~\cite{sun2025visual} targets implicit affordance grounding in egocentric images. 
RSC addresses a complementary gap, exocentric scenario grounding, where the challenge is reasoning over rich paragraph-length scenarios and disambiguating among visually similar instances using rich relational and intentional cues rather than inferring affordances or external knowledge. RSC also provides an OOD split, multi-axis difficulty tags, and per-instance reasoning trace annotations to support diverse learning and analysis. 

Beyond introducing the benchmark, we also propose a strong baseline \textbf{\method}, a two-stage curriculum reasoning method for scenario-based visual grounding. 
In Stage~1, \emph{Thought-Primed SFT (TP-SFT)} aligns the model to the output schema and elicits faithful reasoning traces before a structured answer, using the easier RSC slices to stabilize interface learning. 
In Stage~2, \emph{Incentive-Curriculum GRPO (IC-GRPO)} refines localization and disambiguation via shaped rewards coupling geometry, including smooth IoU reward with center-consistency and out-of-bounds penalties, and alias-aware category rewards. 
The training follows a tag-aware curriculum, feeding more difficult non-unique, cluttered, overlapping, and off-center targets in the later stage.
A prompt-template ensemble further improves robustness across query surface forms. \method\ presents a well-characterized baseline demonstrating that difficulty-aware curriculum training substantially improves scenario grounding and transfers the improvement to standard benchmarks.

Our main contributions are as follows:
\vspace{-0.1in}
\begin{itemize}
    \item We introduce \textbf{RSC}, a scenario-based visual grounding benchmark with difficulty-controlled instances, per-instance reasoning trace annotations, and an OOD split with disjoint object categories.
    \vspace{-0.1in}
    \item We propose \textbf{\method}, a curriculum reasoning method combining supervised warm-starting with difficulty-aware reinforcement learning, providing a strong and well-characterized reference point for scenario-based grounding.
    \vspace{-0.1in}
    \item Experiments demonstrate that RSC exposes failure modes invisible to standard benchmarks, and that difficulty-aware curriculum training transfers improvements to standard referring benchmarks.
\end{itemize}

\begin{figure*}[t!]
    \centering
\includegraphics[width=\linewidth]{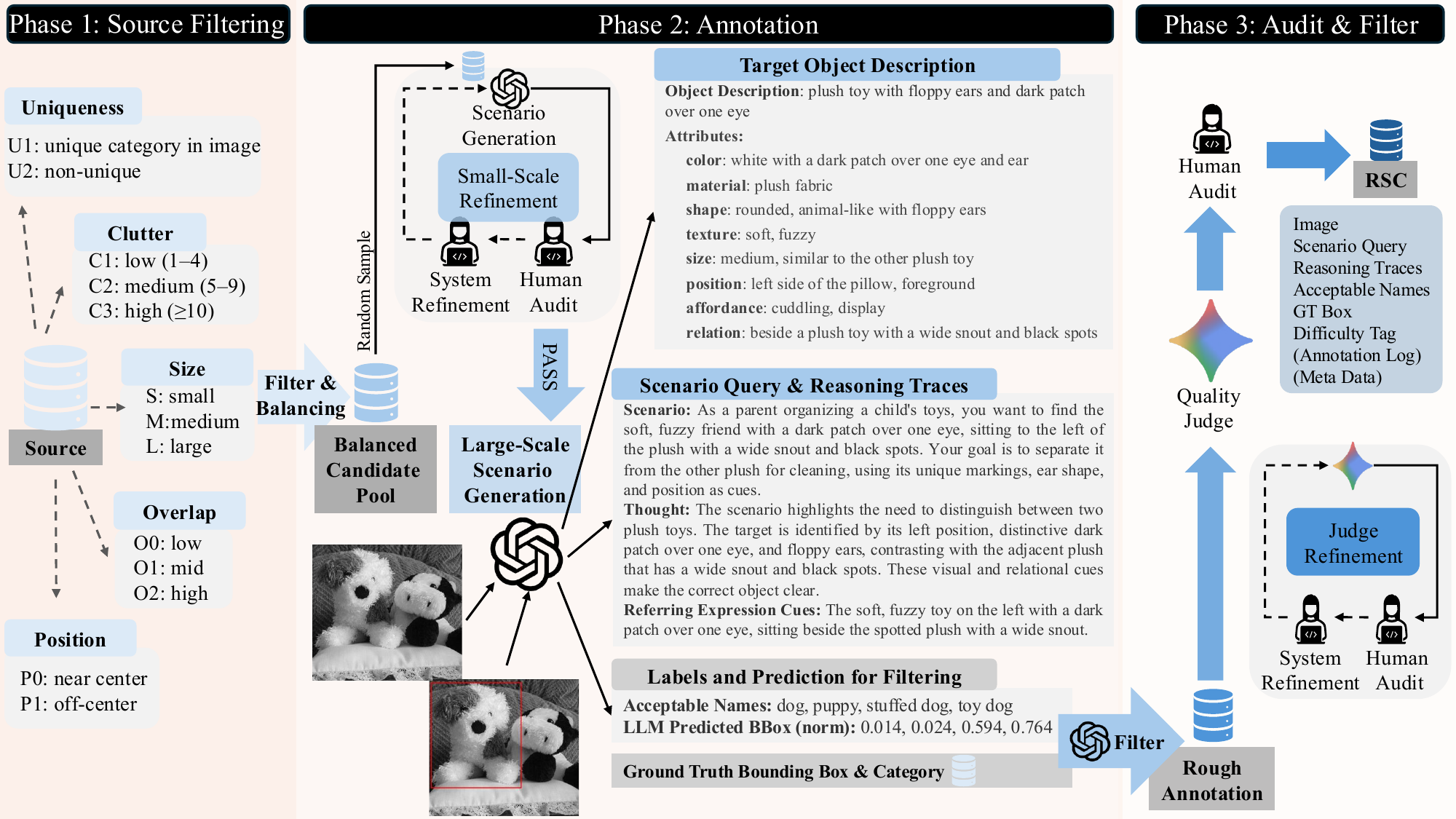}

\caption{
\textbf{Phase 1} filters and balances source instances computing five interpretable difficulty tags to form a tag-balanced candidate pool.
\textbf{Phase 2} generates annotations via a two-stage process: a small-scale refinement loop first validates the generation prompt through iterative system refinement and human audit, before large-scale scenario generation is applied to the full candidate pool.
Each instance is annotated with a target object description, a scenario query with reasoning traces, acceptable category aliases, and an LLM-predicted bounding box for quality filtering.
\textbf{Phase 3} applies automatic and human quality control: a quality judge filters rough annotations, human auditors verify a stratified sample.
The final RSC dataset provides, per instance, a scenario query, reasoning traces, acceptable names, ground-truth box, and difficulty tags.}
\vspace{-0.15in}
\label{fig:rsc-pipeline}
\end{figure*}

\section{Related Work}
\label{sec:related_work}

{\bf Referring Expression Datasets and Methods.} 
Referring expression aims to localize image regions described by natural language. Early datasets~\cite{yu2016refcoco,kazemzadeh2014referitgame,mao2016generation,plummer2015flickr30k} introduced the task using short, literal phrases that directly name the target objects.
Subsequent benchmarks expanded the scope along multiple axes: compositional reasoning with hard negatives~\cite{liu2019clevr,Chen_2020_CVPR,liu2024finecops, dong2026refadv}, structured reasoning over scene graphs~\cite{yang2020graph}, segmentation-based grounding~\cite{lai2024lisa,wu2020phrasecut}, 3D referring expressions~\cite{chen2020scanrefer,achlioptas2020referit3d} and GUI domains~\cite{you2024ferret}. 
More recently, SK-VG~\cite{chen2023advancing} introduces grounding conditioned on long-form scene knowledge, and EgoIntention~\cite{sun2025visual} targets egocentric intention grounding where models must infer the user intent from first-person views.
To the best of our knowledge, no existing benchmark evaluates grounding from scenario-based queries: paragraph-length descriptions specifying user roles, goals, and explicit distractor contrasts, where the challenge lies in reasoning over rich intentional context to disambiguate distractors.

\noindent
{\bf Visual Grounding with Vision-and-Language Models. } 
Early visual grounding methods adopted modular pipelines pairing region proposals with language encoders~\cite{kamath2021mdetr, li2022grounded}. Transformer-based architectures~\cite{deng2021transvg,li2024groundinggpt,zhan2025griffon} later reformulated referring expression as an end-to-end task. Grounding DINO~\cite{liu2024grounding} further unified open-set object detection with text-conditioned localization. With the rise of large vision-language models (LVLMs), grounding has become an inherent capability within general-purpose multimodal systems. Models such as Ferret~\cite{you2023ferret}, Shikra~\cite{chen2023shikra}, and Kosmos-2~\cite{peng2023kosmos} pioneered region-level referring and grounding in LVLMs by representing spatial coordinates as part of the text generation process. Subsequent work scaled these ideas: Qwen-VL~\cite{Qwen2-VL} and Qwen2.5-VL~\cite{Qwen2.5-VL} align image--caption--box tuples during pretraining to support open-vocabulary grounding, while InternVL~\cite{chen2024internvl,chen2024far} and VisionLLM~\cite{wu2024visionllm} extend grounding to hundreds of vision-language tasks. Most recently, UniVG-R1~\cite{bai2025univg} incorporates reinforcement learning with Chain-of-Thought reasoning~\cite{wei2022chain} for universal grounding. 
Despite these advances, existing methods are predominantly evaluated on short object-centric queries, and the scenario-based grounding setting, where success requires reasoning over roles, goals, and relational context, remains unstudied.
RSC is designed to fill this gap, and \method\ provides a strong curriculum reasoning baseline tailored for this setting.

\section{Referring Scenario Comprehension}
\label{sec:rsc}

Referring Scenario Comprehension (RSC) asks a model to localize an object from a \emph{scenario} which describes a user role, a goal, distractors, and non-literal disambiguating cues.
Given an image $x\in\mathbb{R}^{H\times W\times 3}$ and a scenario $s$, a model $\phi$ predicts:
\begin{equation}
\small
\phi(x,s)\;\to\;(\widehat{y},\,\widehat{b}),
\quad
\widehat{b}=(\hat{x},\hat{y},\hat{w},\hat{h})\in\mathbb{Z}_{\ge0}^4,
\label{eq:rsc-task}
\end{equation}
where $\widehat{b}$ is an \texttt{xywh} box clipped to $[0,W]\!\times\![0,H]$.
The dataset provides tuples
\begin{equation}
\small
\mathcal{D}=\big\{(x_i,\,s_i,\,y_i,\,b_i,\,
\mathcal{A}_i,\,e_i,\,r_i,\,\tau_i)\big\}_{i=1}^{N},
\label{eq:rsc-dataset}
\end{equation}
with gold category $y_i$, box $b_i$, acceptable aliases $\mathcal{A}_i$, a concise referring expression $e_i$, a reasoning trace $r_i$, and interpretable difficulty tags $\tau_i=(U_i,C_i,S_i,O_i,P_i)$ for \textbf{U}niqueness, \textbf{C}lutter, \textbf{S}ize, \textbf{O}verlap, and \textbf{P}osition. 
By construction, $s_i$ does not reveal $y_i$. The curation pipeline proceeds in three phases, illustrated in Figure~\ref{fig:rsc-pipeline}.

\vspace{-0.1in}
\paragraph{Phase 1: Source Filtering and Balancing.}
We source instances from MS-COCO~\cite{lin2014microsoft}, preferring images that overlap with RefCOCO/+/g for visual recognizability, and prevent leakage by excluding evaluation image IDs, retaining category labels only as hidden construction signals, and de-duplicating by image hash and annotation ID before splitting. The OOD split draws from LVIS~\cite{gupta2019lvis} with COCO overlaps and synonym collisions removed to ensure strict category disjointness at both string and synset levels.

For each instance $i$ with box $b_i=(x_i,y_i,w_i,h_i)$ and center $c_i$, we compute:
\begin{equation}
\small
a_i=\frac{w_i h_i}{HW}, \
d_i=\frac{\lVert c_i-(\tfrac{W}{2},\tfrac{H}{2})\rVert_2^2}
{W^2+H^2}, \
o_i=\sum_{j\neq i}\mathrm{IoU}(b_i,b_j).
\label{eq:stat-defs}
\end{equation}
Difficulty tags are assigned by quantile binning estimated on the candidate pool \emph{before} splitting: \emph{Size~(S)} via tertiles of $a_i$; \emph{Overlap~(O)} via low/mid/high percentiles of $o_i$; \emph{Position~(P)} via a median split of $d_i$; \emph{Uniqueness~(U)} by the presence of same-category distractors ($m_i{=}0\Rightarrow$U1; $m_i{\ge}1\Rightarrow$U2); and \emph{Clutter~(C)} by binning the per-image instance count $N_{\mathrm{img}}$. 
These axes expose ambiguity~(U), scene density~(C), scale~(S), occlusion and congestion~(O), and off-center placement~(P) in a way that is both interpretable and controllable.

To form a balanced candidate pool, we target split-wise marginal proportions $\Pi=\{\pi_U,\pi_C,\pi_S,\pi_O,\pi_P\}$ and enforce per-category quotas to avoid category dominance. Allocation proceeds hierarchically: (i)~assign a quota per category using long-tail-favoring priorities; (ii)~partition each category's quota across tag combinations according to $\Pi$; (iii)~sample within each bin. A continuous difficulty score $D_i\in[0,1]$ uses a monotone combination of quantile-normalized $(a_i,o_i,d_i)$, distractor count, and category rarity, ordering instances within each bin. 

\vspace{-0.1in}
\paragraph{Phase 2: Annotation.}
Inspired by prior model-in-the-loop pipelines~\cite{kirillov2023segment, wang2023self, xu2304wizardlm}, our scenario generation follows a two-stage process to ensure annotation quality before scaling. 
In a \emph{small-scale refinement loop}, we first generate scenarios for a small set of random samples, then conduct iterative system refinement and a human audit to validate and improve the generation prompt. Generation proceeds to large scale only once the audited pass rate exceeds a quality threshold of 90\%.

In \emph{large-scale generation}, an LLM receives two views of each target image: the full image and the same image with a red rectangle prior marking the target region (the rectangle is not part of the object). It then returns a structured JSON containing: a concise category-free referring expression ($\le$25 words), a user-driven scenario (role, goal, and $\ge$3 disambiguating cues from attributes, relations, position, and affordance), a reasoning trace explaining how the scenario maps to visual evidence, canonical object attributes, an alias set $\mathcal{A}_i$, and a predicted bounding box $\hat{b}_i$. 
The prompt enforces that category names are absent from the scenario and expression, and requires at least one explicit contrast with a plausible distractor.

\begin{table}[t]
\centering
\small
\caption{\textbf{RSC data retention across quality control stages.}
After \emph{LLM GT Filter}: instances passing the automatic schema, box IoU, alias consistency, and leakage gates. After \emph{Quality Filter}: instances passing the full judge-and-audit pipeline. }
\label{tab:data_quality}
\setlength{\tabcolsep}{5pt}
\resizebox{\linewidth}{!}{
\begin{tabular}{lrrcrc}
\toprule
\textbf{Split} & \textbf{Initial} 
  & \textbf{GT Filter} & \textbf{Kept}
  & \textbf{Quality Filter} & \textbf{Kept} \\
\midrule
Train (SFT)  & 30,000 & 29,551 & 98.5\% & 23,802 & 80.5\% \\
Train (RL)   & 10,000 &  9,884 & 98.8\% &  7,540 & 76.3\% \\
Test (ID)    &  5,000 &  4,915 & 98.3\% &  4,038 & 82.2\% \\
Test (OOD)   &  5,000 &  3,983 & 79.7\% &  3,247 & 81.5\% \\
\bottomrule
\end{tabular}}
\vspace{-0.1in}
\end{table}

\vspace{-0.1in}
\paragraph{Phase 3: Audit and Filter.}
Instances passing the automatic gates in Phase~2 form a rough annotation pool, which undergoes dual-track quality control summarized in Table~\ref{tab:data_quality}. The LLM GT filter, comprising schema validation, box IoU gate, alias consistency, and leakage detection, retains 98--99\% of ID instances and 79.7\% of OOD instances. 
The lower OOD retention reflects the greater visual ambiguity of LVIS categories, for which the LLM more frequently fails to produce an accurate box or alias set. This confirms the filter is performing meaningful quality control on harder categories rather than passing annotations indiscriminately.

Instances passing the LLM GT filter then enter the quality filter stage. A \emph{quality judge} automatically scores each annotation for the uniqueness of the scenario referring target, and the accuracy of the bounding box for the scenario. Borderline cases enter a \emph{judge refinement} loop applying targeted system and human correction before re-scoring. After this stage, 76--82\% of LLM-filtered instances are retained across splits, yielding final counts of 23,802 training (SFT), 7,540 training (RL), 4,038 in-domain test, and 3,247 OOD test examples.

In parallel, three independently drawn samples of 100 instances each undergo \emph{human audit} by three expert annotators, who independently verify scenario non-leakage, alias consistency ($y_i\in\mathcal{A}_i$), box correctness ($\mathrm{IoU}(\hat{b}_i,b_i)\ge\theta_{\mathrm{bbox}}$), and attribute faithfulness. 
Majority-vote accuracy was 95.7\% overall (per-rater: 94\%/96\%/97\%), with substantial agreement (Fleiss’ $\kappa{=}\,$\textit{0.94}). The residual 4\% errors were attributable to  LLM-hallucinated distractor descriptions (${\sim}3\%$) and minor attribute misdescriptions (${\sim}1\%$). 

Each released RSC instance provides: a scenario query, reasoning traces, acceptable category names, ground-truth bounding box, and difficulty tag, with annotation logs and metadata included for research use. \textit{More details are provided in the appendix.}

\begin{figure}[t]
    \centering
    \includegraphics[width=\linewidth]{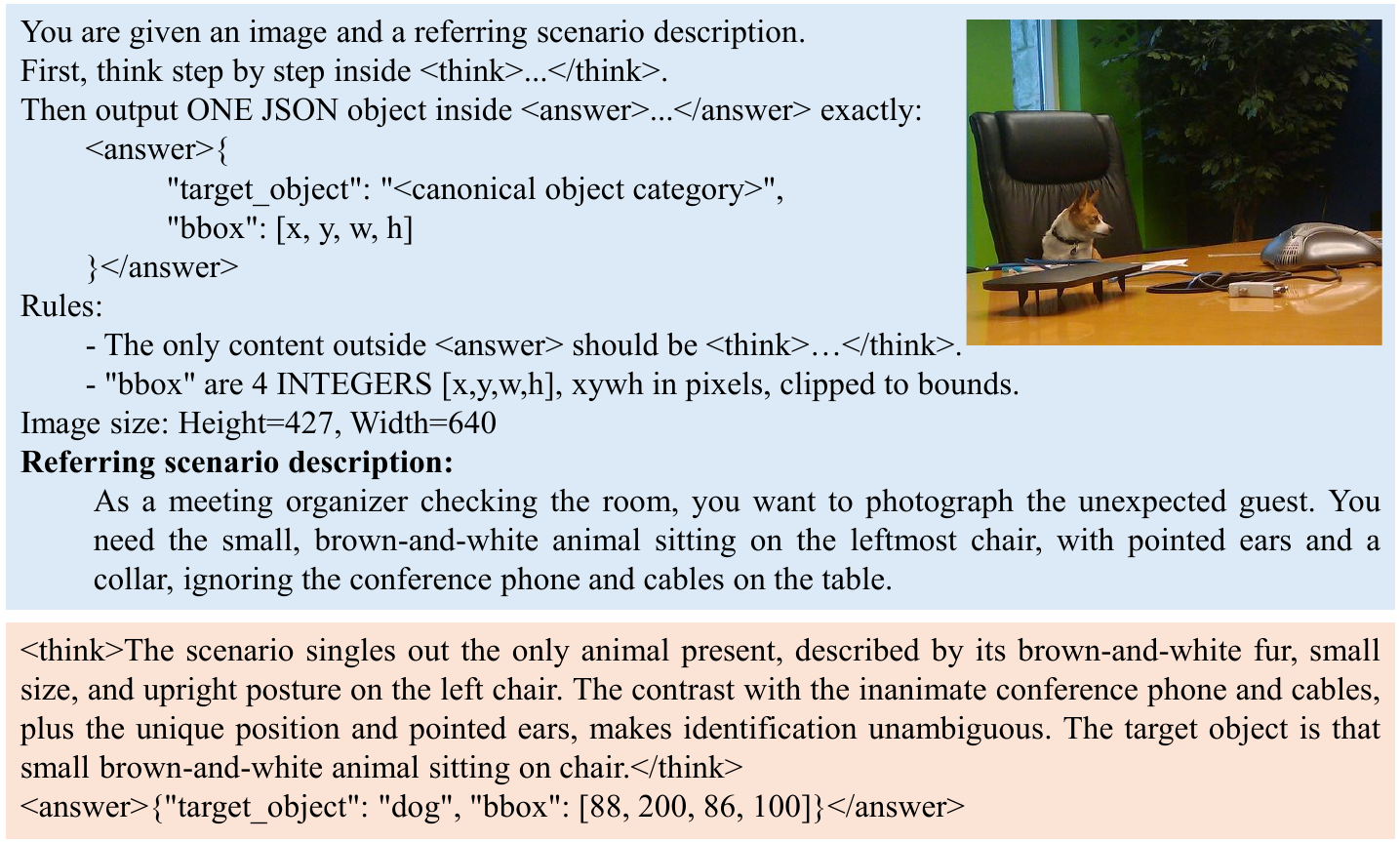}
    \label{fig:\method-schema}
    \caption{\textbf{\method\ prompt and output schema.} Given an image and a user-driven scenario, the model is instructed to reason inside \texttt{\textless think\textgreater} and to emit a structured JSON inside \texttt{\textless answer\textgreater}. The JSON contains \texttt{target\_object} and \texttt{bbox} \texttt{[x,y,w,h]}.
    Scenarios avoid category names and force disambiguation via attributes, relations, and spatial cues, while \texttt{\textless think\textgreater} must justify the selection and ignore distractors. This schema is used in both TP-SFT and IC-GRPO. To increase the robustness of \method, IC-GRPO uses 7 other templates.
    }
    \vspace{-0.2in}
\end{figure}

\section{\method}
\label{sec:method}

\method\ is a two-stage curriculum reasoning method for scenario-based grounding on RSC. Thought-Primed SFT (TP-SFT) aligns the interface and elicits faithful \texttt{<think>} traces that precede \texttt{<answer>}. Incentive-Curriculum GRPO (IC-GRPO) then refines reasoning and localization by optimizing shaped rewards over RSC's difficulty-stratified curriculum.

\subsection{Thought-Primed SFT}

Given image $x$, scenario $s$, and target text $\mathbf{y}$ (the concatenation of \texttt{<think>} and \texttt{<answer>} spans), we optimize the standard next-token loss:
\begin{equation}
\small
\mathcal{L}_{\mathrm{SFT}}(\theta)
= - \mathbb{E}_{(x,s,\mathbf{y})\sim\mathcal{D}_{\mathrm{sft}}}
\left[\sum_{t=1}^{|\mathbf{y}|}
\log p_\theta\!\left(y_t \mid x,s,\mathbf{y}_{<t}\right)\right].
\end{equation}
The output schema requires a single JSON inside \texttt{<answer>} with keys \texttt{target\_object} and \texttt{bbox} (xywh, pixel integers). Training uses the easier RSC slices ($D_i$ easy percentiles) to stabilize schema learning before RL. TP-SFT teaches the output schema, elicits faithful reasoning traces inside \texttt{<think>}, and provides a stable reference policy $\pi_{\mathrm{ref}}$ for the subsequent RL stage.

\subsection{Incentive-Curriculum GRPO}

We fine-tune $\pi_\theta$ using GRPO~\cite{guo2025deepseek} with group-relative advantages (Eq.~\ref{eq:advantage}) and a KL-regularized objective with adaptive $\beta$ tracking a target KL band throughout training (see Appendix~\ref{app:method}).

\paragraph{Shaped rewards.}
The scalar reward combines four components (Eq.~\ref{eq:rtotal}). The \emph{geometry reward} $r_{\mathrm{iou}}$ combines a base IoU term with smooth logistic bonuses near two operating points, a small center-consistency term, and a penalty for out-of-bounds predictions. The \emph{category reward} $r_{\mathrm{cat}}$ is alias-aware: it awards full credit for canonical names, partial credit for accepted aliases and token-overlap matches, and is gated by a minimum IoU threshold to discourage well-labelled but poorly localized predictions. The \emph{format} and \emph{structure rewards} enforce schema compliance, rewarding parseable JSON inside \texttt{<answer>} with the required keys and penalizing malformed outputs. Reward weights are linearly annealed to increase geometry emphasis in later training.

\vspace{-0.1in}
\paragraph{Tag-aware curriculum.}
IC-GRPO samples from RSC using difficulty scores $D_i$. Stage~1 draws predominantly from easy-to-medium slices; Stage~2 shifts toward harder instances with non-unique categories~(U2), high clutter~(C3), high overlap~(O2), and off-center placement~(P1). This progressive schedule addresses reward sparsity: easy instances first establish reliable IoU signals before the policy encounters harder disambiguations where the category reward requires a cleared IoU gate.

\vspace{-0.1in}
\paragraph{Prompt-template ensemble.}
To reduce sensitivity to surface-level query phrasing, 
we uniformly sample from eight prompt paraphrases 
(PTE-8) per training step. All templates share the same 
output schema; rewards are logged per template for 
analysis but supervision is identical, improving 
robustness without changing the learning objective.

\begin{table*}[t]
\centering
\footnotesize
\renewcommand{\arraystretch}{1.2}
\setlength{\tabcolsep}{4pt}
\caption{\textbf{Performance on RSC} (ID and OOD). Metrics: mIoU and $\mathrm{Acc}@\{0.5,0.7\}$ (higher is better); Cat Acc = category accuracy.
$^\ddagger$\textbf{Oracle settings, not directly comparable:} Grounding DINO receives privileged inputs unavailable 
at inference: 
\emph{cat token} provides the gold category name directly and \emph{ref.\ cue} feeds the conciser reasoning trace conclusion "referring expression cue" (See Figure~\ref{fig:rsc-pipeline}) as the open-vocabulary query.}
\label{tab:rsc_main}
\begin{tabular}{lcccccccc}
\toprule
& \multicolumn{4}{c}{RSC-ID} & \multicolumn{4}{c}{RSC-OOD} \\
\cmidrule(lr){2-5}\cmidrule(lr){6-9}
Model & mIoU & Acc@0.5 & Acc@0.7 & Cat Acc & mIoU & Acc@0.5 & Acc@0.7 & Cat Acc \\
\midrule
GPT-4o~\cite{hurst2024gpt}           & 19.41 & 13.23 &  5.37 & 79.45 & 16.57 &  9.55 & 3.08 & \textbf{62.00} \\
Claude 3.7~\cite{anthropic2025claude37}       & 16.64 & 8.32 & 3.71 & 89.67 & 12.04 & 5.54 & 1.87 & 58.98 \\
\midrule
Grounding DINO (\textit{cat token})$^\ddagger$   & 44.60 & 47.55 & 42.03 & \textemdash & 32.18 & 31.99 & 27.89 & \textemdash \\
Grounding DINO (\textit{ref. cue})$^\ddagger$   & 48.99 & 51.84 & 46.02 & \textemdash & 38.12 & 38.26 & 34.07 & \textemdash \\
\addlinespace[4pt]
\hdashline
\addlinespace[4pt]
InternVL2.5 8B~\cite{chen2024expanding}      & 16.76 & 11.88 & 6.74 & 81.70 & 8.08 & 3.64 & 1.61 & 36.50 \\
Qwen3-VL 8B~\cite{qwen2025qwen3vl}        & 15.46 & 11.17 & 6.05 & 75.04 & 7.38 & 3.70 & 1.48 & 46.97 \\
Qwen2.5-VL 7B~\cite{Qwen2.5-VL}    & 30.31 & 27.42 & 15.66 & 30.86 & 21.54 & 15.88 & 9.19 & 20.82 \\
\midrule
\method\ (Ours)    & 55.68 & 60.90 & 42.32 & 94.23 & 38.37 & 38.11 & 22.64 & 21.13 \\
\bottomrule
\end{tabular}
\end{table*}

\begin{table*}[t]
\centering
\footnotesize
\renewcommand{\arraystretch}{1.2}
\setlength{\tabcolsep}{5pt}
\caption{\textbf{Ablation of curriculum in IC-GRPO} on RSC in-domain (RSC-ID) and out-of-domain (RSC-OOD).
Metrics: mIoU and $\mathrm{Acc}@\{0.5,0.7\}$ for boxes; Cat Acc for category naming.
\emph{Single Stage} trains on the union of all RL samples without curriculum; \emph{Stage-$k$} trains on the stage-$k$ slice, all other settings unchanged.}
\label{tab:rsc_ablation_curriculum}
\begin{tabular}{lcccccccc}
\toprule
& \multicolumn{4}{c}{RSC-ID} & \multicolumn{4}{c}{RSC-OOD} \\
\cmidrule(lr){2-5}\cmidrule(lr){6-9}
Method & mIoU & Acc@0.5 & Acc@0.7 & Cat Acc & mIoU & Acc@0.5 & Acc@0.7 & Cat Acc \\
\midrule
Qwen2.5-VL 7B~\cite{Qwen2.5-VL} & 30.31 & 27.42 & 15.66 & 30.86 & 21.54 & 15.88 & 9.19 & 20.82 \\
SFT                         & 55.01 & 60.51 & 42.59 & 89.05 & 33.20 & 34.03 & 20.36 & 12.53 \\
GRPO, Single Stage          & 54.04 & 59.04 & 38.43 & 94.90 & 37.66 & 36.99 & 20.82 & 20.05 \\
GRPO, Stage 1               & \textbf{55.68} & 60.95 & 41.88 & 94.04 & 36.93 & 36.59 & 21.37 & 19.00 \\
GRPO, Stage 2               & \textbf{55.68} & 60.90 & \textbf{42.32} & \textbf{94.23} & \textbf{38.37} & \textbf{38.11} & \textbf{22.64} & \textbf{21.13} \\

\bottomrule
\end{tabular}%
\vspace{-0.1in}
\end{table*}

\section{Experimental Setup}
\label{sec:exp_setup}

\paragraph{Implementation details.}
Annotation is generated by GPT4o~\cite{hurst2024gpt} and the quality judge relies on Gemini-2.5-Pro~\cite{comanici2025gemini}.
We train \method\ on Qwen2.5-VL-7B-Instruct~\cite{Qwen2.5-VL} 
in two stages. TP-SFT uses AdamW with learning rate 
$5{\times}10^{-6}$ for 5 epochs. GRPO Stage~1 uses AdamW 
(lr $1{\times}10^{-6}$) for 5 epochs with $K{=}6$ rollouts 
per prompt; Stage~2 uses lr $2{\times}10^{-6}$ for 2 epochs 
with $K{=}12$ rollouts, generation batch 288, temperature 
$0.9$, top-$p$ $0.92$, and max completion length 160. 
The KL coefficient $\beta$ is initialized at 
$2{\times}10^{-2}$ and adapted online in both stages. 
All remaining hyperparameters are 
provided in Appendix~\ref{app:exp_setup}.

\vspace{-0.1in}
\paragraph{Data for curriculum learning.}
We partition the raw instances at the image level into disjoint sets $\mathcal{S}_{\mathrm{SFT}}$, $\mathcal{S}_{\mathrm{RL}}$, and $\mathcal{S}_{\mathrm{test}}$, enforcing no image leakage across splits. After RSC curation filters, we retain 23k instances for SFT and 7k for RL. Difficulty is binned by global quantiles of $D_i$ into easy, medium, and hard buckets. The SFT split targets an easy-purity of at least 0.70. RL Stage~1 uses mixture $\Pi^{(1)}{=}(0.70,\,0.30,\,0.00)$ and Stage~2 shifts to $\Pi^{(2)}{=}(0.20,\,0.60,\,0.20)$ for (easy, medium, hard).

\section{Experiments}
\label{sec:exp_results}
\begin{table}[t]
\renewcommand{\arraystretch}{1.2}  %
\footnotesize
\centering
\setlength{\tabcolsep}{2.5pt}
\caption{\textbf{Referring expression comprehension} on RefCOCO+ and RefCOCOg.
Metric: Acc@0.5 (\%, $\uparrow$).
$^\dagger$ Use the official grounding pipeline and prompt~\cite{Qwen2.5-VL} .
$^\ast$Use a custom prompt.}
\label{tab:refcoco_main}
\begin{tabular}{lccccc}
\toprule
& \multicolumn{3}{c}{RefCOCO+} & \multicolumn{2}{c}{RefCOCOg} \\
\cmidrule(lr){2-4}\cmidrule(lr){5-6}
Method & Val & Test A & Test B & Val & Test \\
\midrule
Qwen2.5\text{-}VL     & 84.20 & 89.10 & 76.90 & 87.20 & 87.20 \\
Qwen2.5\text{-}VL$^\ast$             & 52.54 & 56.75 & 52.53 & 52.46 & 51.40 \\
\method$^\ast$, SFT                                 & 50.86 & 43.84 & 50.74 & 65.45 & 63.39 \\
\method$^\ast$, GRPO                                & 70.16 & 74.63 & 70.05 & 78.19 & 75.61 \\
\bottomrule
\end{tabular}%
\vspace{-0.1in}
\end{table}

\paragraph{Comparison on Referring Scenario Comprehension.}
Table~\ref{tab:rsc_main} evaluates two closed-source LLMs~\cite{hurst2024gpt,anthropic2025claude37}, one open-vocabulary grounding model~\cite{liu2024grounding}, three open-source LVLMs~\cite{Qwen2.5-VL,qwen2025qwen3vl,chen2024expanding} on RSC. The results reveal a consistent pattern across baselines: models with strong category accuracy tend to lag on localization, while strong detectors lack semantic reasoning. This trade-off reflects the core challenge RSC is designed to expose: scenario grounding requires jointly strong spatial localization and scenario-aware semantic inference, a combination underemphasized in prior benchmarks.

The Grounding DINO results illustrate the gap between detection-style grounding and full scenario reasoning. Under the oracle \emph{cat token} setting, which directly supplies the gold category name, Grounding DINO achieves strong localization. Under the \emph{ref.\ cue} setting, which supplies a short descriptive non-literal phrase rather than the full scenario, performance improves further, particularly on OOD. These gains confirm that Grounding DINO's architecture is well-optimized for direct-cue detection, but neither setting reflects the full RSC protocol given to other models.

Among models that receive only the full scenario query $s$ and must jointly predict the target category and bounding box, \method\ substantially outperforms all baselines on ID mIoU. Closed-source models achieve high category accuracy but poor localization, confirming that understanding a scenario does not automatically translate to spatial grounding. The localization and target category prediction gains from \method\ are consistent across ID and OOD splits, reducing the baseline trade-off and delivering robust performance on the full RSC protocol.

\vspace{-0.1in}
\paragraph{Effectiveness of GRPO and Curriculum Learning.}
Table~\ref{tab:rsc_ablation_curriculum} ablates training stages. TP-SFT already delivers a large ID jump over the off-the-shelf VLM~\cite{Qwen2.5-VL}, showing that aligning the model with the reasoning schema and the RSC objective teaches strong box prediction. However, OOD category accuracy remains the dominant failure mode after SFT.
Moving to IC-GRPO without curriculum (\emph{Single Stage}) improves OOD across both localization and category naming, but slightly regresses ID localization relative to SFT. This is consistent with reward sparsity: when easy and hard cases are mixed, many early trajectories fall below the IoU gate that modulates the alias reward, yielding high-variance gradients and modest gains.

The tag-aware curriculum addresses this. \emph{Stage~1} (easy to medium) preserves the strong ID accuracy of SFT and improves OOD over SFT, but falls short of Stage~2. 
\emph{Stage~2} (medium to hard) yields the best results across all OOD metrics while keeping ID performance essentially unchanged. 
As geometric accuracy improves on harder, more ambiguous instances, the policy learns to map scenario cues to both \emph{where} and \emph{what} under category shift. 
However, ScenGround's OOD category accuracy remains only marginally above the untuned baseline, indicating that cross-category semantic naming is largely unsolved by curriculum training alone. This may be attributed to the fine-grained LVIS categories in the OOD split (see Appendix~\ref{app:rsc_stats}), and suggests the need for stronger semantic generalization or open-vocabulary training strategies. Overall, SFT establishes the interface and ID competence, while IC-GRPO with a difficulty-aware curriculum closes the OOD localization gap.

\vspace{-0.15in}
\paragraph{Comparisons on standard referring-expression 
benchmarks.}
Table~\ref{tab:refcoco_main} reports Acc@0.5 on RefCOCO+~\cite{yu2016refcoco} and RefCOCOg~\cite{mao2016generation}.
The first row (Qwen2.5-VL$^\dagger$) reflects the model's official phrase-grounding pipeline and prompt~\cite{Qwen2.5-VL}. 
To standardize interfaces for cross-model comparison, we also evaluate Qwen2.5-VL under a custom prompt (row~2, $^\ast$); this prompt departs from its native detection-style prompt and substantially lowers accuracy, illustrating the sensitivity of phrase-grounding systems to prompt and decoding design.

Under this custom prompt, \method\ shows two clear trends. First, SFT underperforms on RefCOCO+ but is notably stronger on RefCOCOg, consistent with RefCOCOg's longer, more descriptive phrases being closer to RSC's scenario style. 
Second, IC-GRPO yields large, consistent improvements across all splits, narrowing the gap to task-specialized pipelines. This suggests that geometry and schema rewards, together with the tag-aware curriculum, help the model resolve intra-class ambiguity and attend to disambiguating cues that typical phrase-grounding prompts would otherwise supply explicitly. Overall, while \method\ is not optimized for RefCOCO-style prompting, its GRPO stage transfers robustly to these benchmarks, suggesting that curriculum training develops transferable reasoning skills rather than overfitting to a specific prompt template.

\vspace{-0.1in}
\paragraph{Analysis on localization ability.}
Table~\ref{tab:rsc_localization_improve} isolates localization by asking models to predict a box only, without category prediction. Qwen2.5-VL's box-only performance closely mirrors its full-task localization, suggesting that category prediction is not its primary bottleneck. By contrast, \method\ delivers strong OOD localization in the box-only regime, substantially outperforming Qwen2.5-VL and confirming that it can ground scenarios from visual cues even when the target category is unseen. The small gap between \method's box-only and full-task results further implies that most remaining OOD degradation comes from naming rather than spatial grounding. Notably, SFT is marginally stronger than GRPO in the box-only setting, while GRPO surpasses SFT on the full task. This suggests that IC-GRPO reallocates capacity to couple localization with scenario understanding rather than optimizing pure box regression.

\begin{table}[t]
\centering
\renewcommand{\arraystretch}{1.4}  %
\fontsize{8.5}{9}\selectfont
\setlength{\tabcolsep}{2.5pt}
\caption{Localization under the \emph{box-only} setting on RSC in-distribution (RSC-ID) and out-of-distribution (RSC-OOD).
Metrics are mIoU and $\mathrm{Acc}@\{0.5,0.7\}$.}
\label{tab:rsc_localization_improve}
\begin{tabular}{lcccccc}
\toprule
& \multicolumn{3}{c}{RSC-ID} & \multicolumn{3}{c}{RSC-OOD} \\
\cmidrule(lr){2-4}\cmidrule(lr){5-7}
Method & mIoU & \shortstack{Acc \\ @0.5}& \shortstack{Acc \\ @0.7} & mIoU & \shortstack{Acc \\ @0.5} & \shortstack{Acc \\ @0.7} \\
\midrule
Qwen2.5-VL 7B  & 27.37 & 24.09 & 13.42 & 19.11 & 13.50 & 7.70 \\
\method, SFT                            & 51.18 & 55.80 & 37.99 & 40.71 & 42.65 & 25.86 \\
\method, GRPO                           & 50.90 & 54.38 & 37.01 & 40.11 & 40.30 & 24.50 \\
\bottomrule
\end{tabular}%
\vspace{-0.1in}
\end{table}

\vspace{-0.1in}
\paragraph{Analysis on reasoning ability.}
Table~\ref{tab:rsc_reasoning_improve} analyzes category prediction under three settings: \emph{Text Only} (scenario $s$ only), \emph{Text-Image} (scenario + image, no box), and \emph{Standard} (full task). 
Removing the localization requirement substantially boosts Qwen2.5-VL on RSC-ID but yields a smaller gain on RSC-OOD, showing that its category decisions are sensitive to the localization constraint. By contrast, \method\ exhibits a small gap between \emph{Text-Image} and \emph{Standard}, indicating that it can jointly localize and classify with minimal degradation. Notably, SFT's strong ID-category accuracy comes at the cost of OOD text-only generalization, while GRPO closes this gap, suggesting that RL improves scenario understanding beyond what schema alignment alone provides. 
Finally, \emph{Text-Image} outperforms \emph{Text Only} across all models and splits, confirming that visual evidence materially helps resolve scenario ambiguities and that models are not relying solely on language priors.

\begin{table}[t]
\fontsize{8.5}{9}\selectfont
\centering
\renewcommand{\arraystretch}{1.4}  %
\setlength{\tabcolsep}{2.5pt}
\caption{Category accuracy on RSC in-distribution (ID) and out-of-distribution (OOD) under three input–output settings.
\emph{Text Only}: predict category from scenario only.
\emph{Text-Image}: predict category from scenario+image (no box).
\emph{Standard}: full task.}
\label{tab:rsc_reasoning_improve}
\begin{tabular}{lcccccc}
\toprule
& \multicolumn{2}{c}{Text Only} & \multicolumn{2}{c}{Text-Image} & \multicolumn{2}{c}{Standard} \\
\cmidrule(lr){2-3}\cmidrule(lr){4-5}\cmidrule(lr){6-7}
Method & ID & OOD & ID & OOD & ID & OOD \\
\midrule
Qwen2.5-VL 7B   & 56.24 & 22.34 & 77.54 & 33.33 & 29.65 & 19.28 \\
\method-SFT     & 81.03 & 10.69 & 93.66 & 14.92 & 92.72 & 12.30 \\
\method-GRPO    & 85.13 & 20.22 & 94.73 & 22.68 & 93.71 & 20.67 \\
\bottomrule
\end{tabular}%
\end{table}

\begin{figure}[t]
    \centering
    \includegraphics[width=\linewidth]{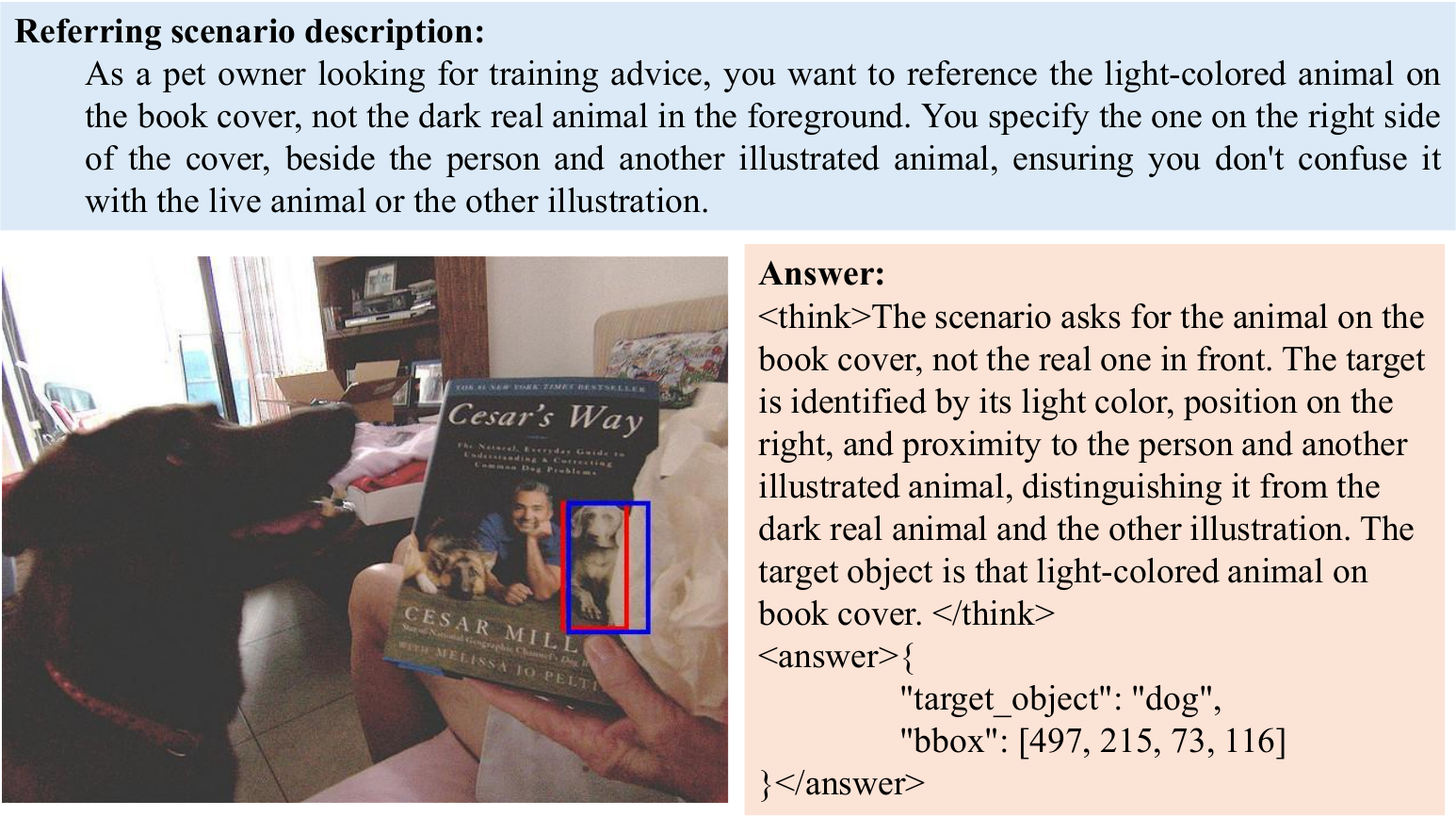}
    \caption{\textbf{Qualitative results on RSC.} Red box for ground truth and blue box for \method\ prediction.}
    \label{fig:qualitative}
    \vspace{-0.2in}
\end{figure}

\paragraph{Qualitative results.}
Figure~\ref{fig:qualitative} illustrates how \method\ grounds user-driven scenarios. 
In the shown example, \method\ correctly distinguishes the \emph{illustrated} animal on the book cover from the real dog by leveraging the cues. 
\textit{More qualitative examples are in the Appendix~\ref{sec:qualitative_examples_supp}. }

\section{Conclusion}
\label{sec:conclusion}

We introduce RSC, a benchmark for scenario-based visual grounding where targets are identified from paragraph-length queries specifying user roles, goals, and explicit distractor contrasts. 
RSC provides per-instance difficulty tags, paired reasoning traces, and an OOD split. Evaluations reveal that strong performance on standard referring-expression benchmarks does not transfer to scenario-based queries, exposing a systematic failure mode invisible to prior evaluation protocols. We further proposed \method, a curriculum reasoning baseline combining supervised warm-starting with difficulty-aware reinforcement learning, which substantially reduces the localization--semantics trade-off on RSC.
We hope RSC provides a useful testbed for studying referential reasoning, with natural extensions toward multi-object, temporal, and interactive grounding settings.

\bibliography{custom}

\appendix

In this appendix, we provide
statistical details of RSC in Section~\ref{app:rsc_stats},
RSC data curation details in Section~\ref{app:rsc_curation},
RSC data examples in Section~\ref{sec:data_examples_supp},
more details of \method\ in Section~\ref{app:method},
tag-level anlysis in Section~\ref{app:tag},
\method\ qualitative examples in Section~\ref{sec:qualitative_examples_supp},
training details in Section~\ref{app:exp_setup},
limitations and future work in Section~\ref{app:limitations},
and broader impact in Section~\ref{app:broader_impact}.

\section{RSC Statistics}
\label{app:rsc_stats}

\begin{figure}[t]
\centering
\includegraphics[width=\linewidth]{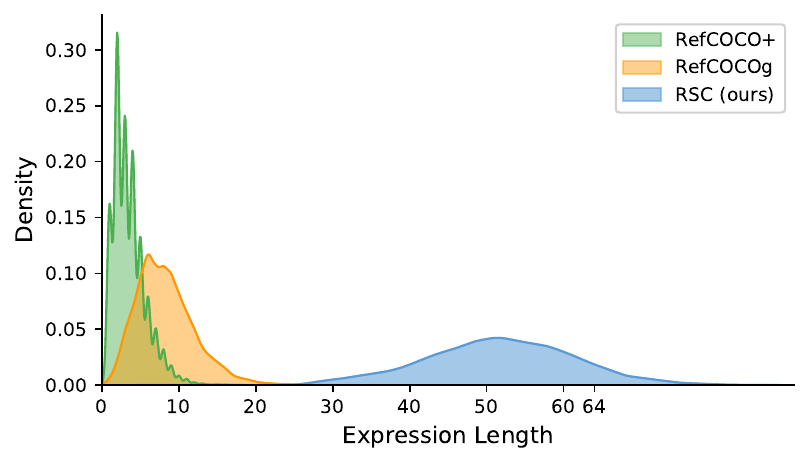}
\caption{\textbf{Query length distribution.} RSC queries are substantially longer than RefCOCO+ and RefCOCOg expressions, reflecting the paragraph-length scenario descriptions that specify user roles, goals, and disambiguating cues. RefCOCO+ and RefCOCOg peak below 10 words; RSC peaks around 50--60 words.}
\label{fig:expr_length_dist}
\end{figure}

\begin{figure}[t]
\centering
\includegraphics[width=\linewidth]{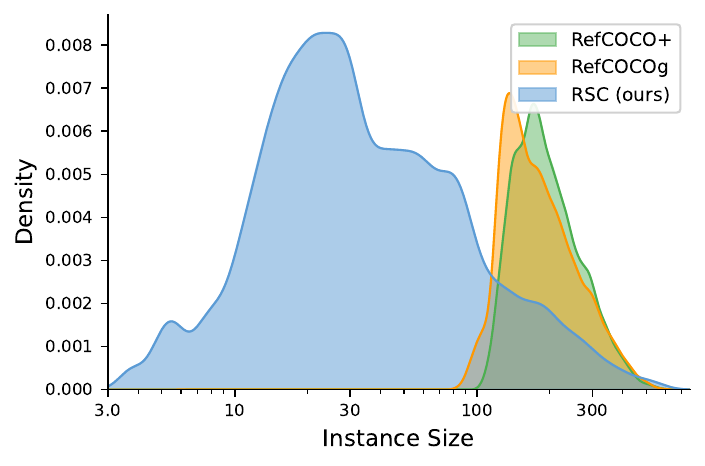}
\caption{\textbf{Instance size distribution.} RSC covers a broader range of instance sizes than RefCOCO+ and RefCOCOg, with notably higher density at smaller instances.}
\label{fig:instance_size_dist}
\end{figure}

\paragraph{Comparisons to RefCOCO+ and RefCOCOg.} We compare the query length and instance size distributions.
Figure~\ref{fig:expr_length_dist} compares query length distributions across RSC, RefCOCO+, and RefCOCOg, illustrating that RSC's scenario-based queries are substantially longer than the short literal phrases used in prior benchmarks. 
Figure~\ref{fig:instance_size_dist} compares instance size distributions, showing that RSC covers a broader range of target scales. Together, these distributions reflect RSC's design goals: longer queries force models to reason over rich contextual descriptions rather than match salient keywords, while the wider size range stresses localization under various controlled conditions.
Table~\ref{tab:vocab_size} further shows that RSC's unique query vocabulary (9,086 tokens) is roughly twice that of RefCOCO+ and RefCOCOg, confirming that the length difference reflects genuine linguistic diversity rather than repetitive padding.

\begin{table}[t]
\centering
\small
\caption{\textbf{Vocabulary size comparison.} Number of unique words across all queries in each dataset's test split.}
\label{tab:vocab_size}
\begin{tabular}{lc}
\toprule
\textbf{Dataset} & \textbf{Unique Vocabulary} \\
\midrule
RefCOCO+  & 4,240 \\
RefCOCOg  & 4,917 \\
RSC (Ours) & \textbf{9,086} \\
\bottomrule
\end{tabular}
\end{table}

\begin{figure*}[t]
\centering
\includegraphics[width=\linewidth]{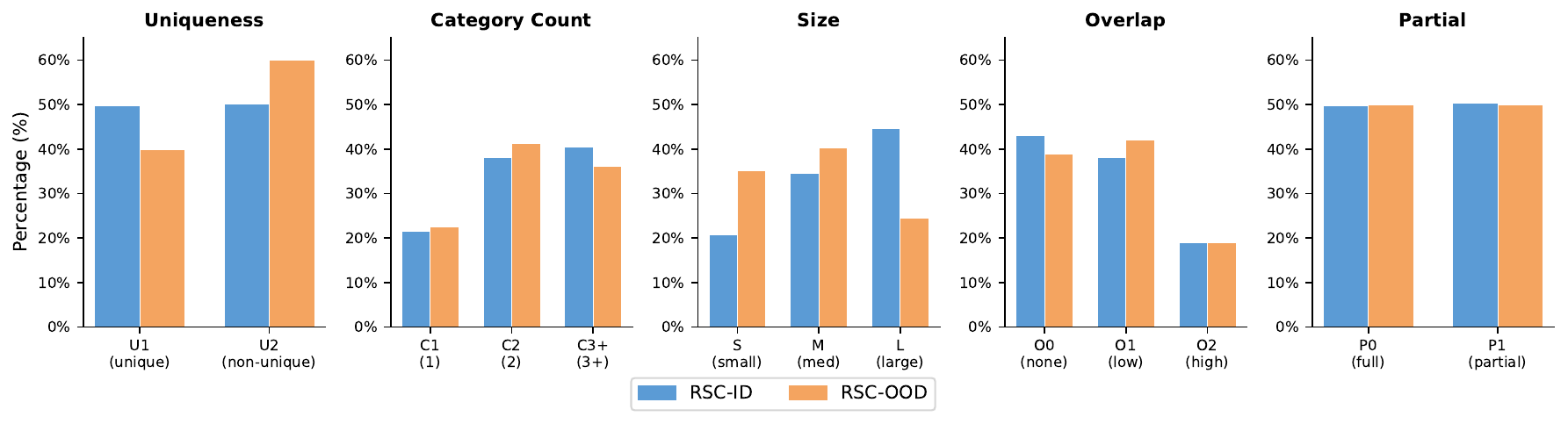}
\caption{\textbf{Difficulty tag distributions for RSC-ID and RSC-OOD.} Each panel shows the per-tag marginal distribution for the in-distribution (RSC-ID) and out-of-distribution (RSC-OOD) test splits across all five axes: Uniqueness~(U), Clutter~(C), Size~(S), Overlap~(O), and Position~(P).}
\label{fig:tag_distributions}
\end{figure*}

\paragraph{RSC difficulty tag distribution.}
Figure~\ref{fig:tag_distributions} shows the difficulty tag distributions for RSC-ID and RSC-OOD test splits. The ID split maintains near-balanced marginals across all five axes, reflecting the tag-balanced sampling strategy described in Section~\ref{sec:rsc}. 
The OOD split shows a notable skew toward non-unique~(U2) and smaller~(S/M) instances, which partially explains the consistently lower accuracy on the OOD split. Distributions across Clutter~(C), Overlap~(O), and Position~(P) remain broadly comparable between ID and OOD, confirming that the primary challenge of the OOD split is category shift rather than a confounding change in scene difficulty along other axes.

\begin{figure*}[t]
\centering
\includegraphics[width=\linewidth]{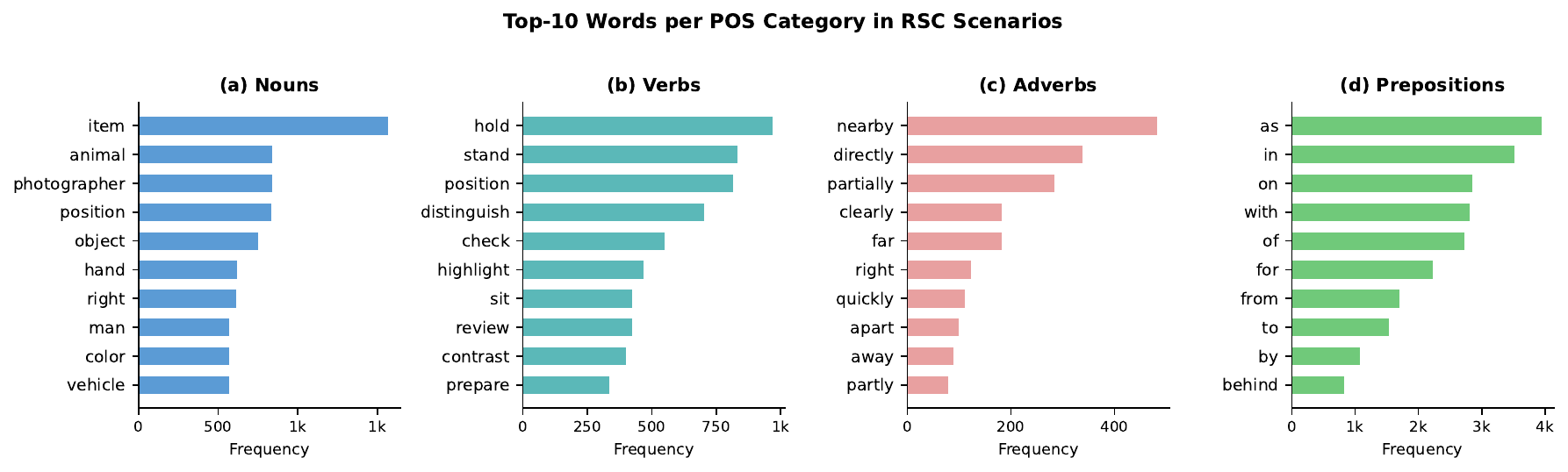}
\caption{\textbf{Top-10 words per part-of-speech category in RSC scenarios.} Frequency counts of the most common nouns, verbs, adverbs, and prepositions across all RSC scenario queries.}
\label{fig:pos_frequency}
\end{figure*}

\paragraph{RSC scenario linguistic analysis.}
Figure~\ref{fig:pos_frequency} shows the top-10 words per part-of-speech category across RSC scenario queries. The noun distribution is dominated by general referential terms (\emph{item}, \emph{object}) and scene participants (\emph{man}, \emph{animal}, \emph{photographer}), confirming that scenarios describe roles and contexts rather than naming target categories directly. The verb distribution reflects the action- and goal-oriented nature of scenario queries, with terms such as \emph{distinguish}, \emph{contrast}, \emph{highlight}, and \emph{review} indicating that queries explicitly require disambiguation reasoning. The preposition distribution is rich in spatial terms (\emph{on}, \emph{behind}, \emph{from}, \emph{by}), consistent with RSC's emphasis on relational and positional cues for grounding. Together, these distributions support the claim that RSC queries demand a qualitatively different form of language understanding than the short attribute-and-location phrase queries in standard REC benchmarks.

\begin{figure}[t]
\centering
\small
\caption{\textbf{Top 20 most frequent categories} in RSC-ID and RSC-OOD test splits.}
\label{tab:category_top20}
\includegraphics[width=\linewidth]{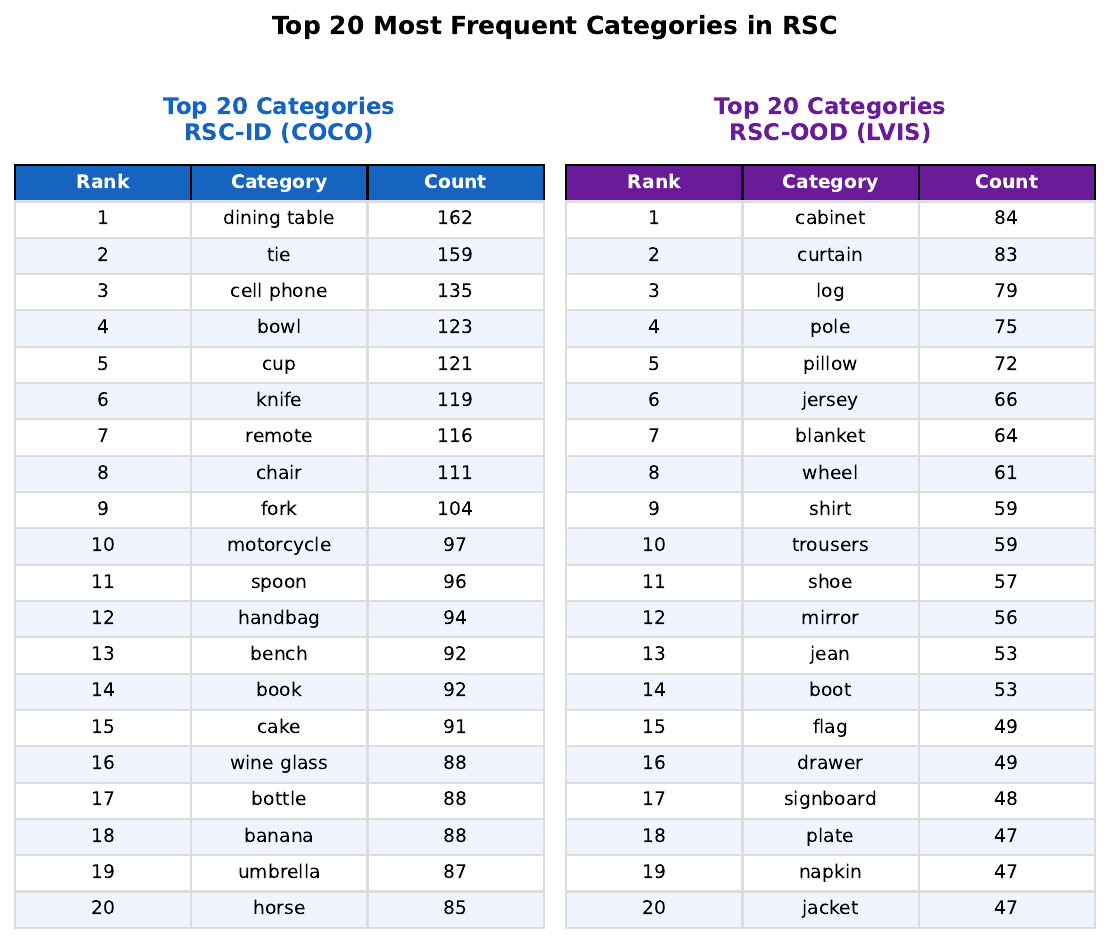}
\end{figure}

\begin{figure*}[t]
\centering
\includegraphics[width=\linewidth]{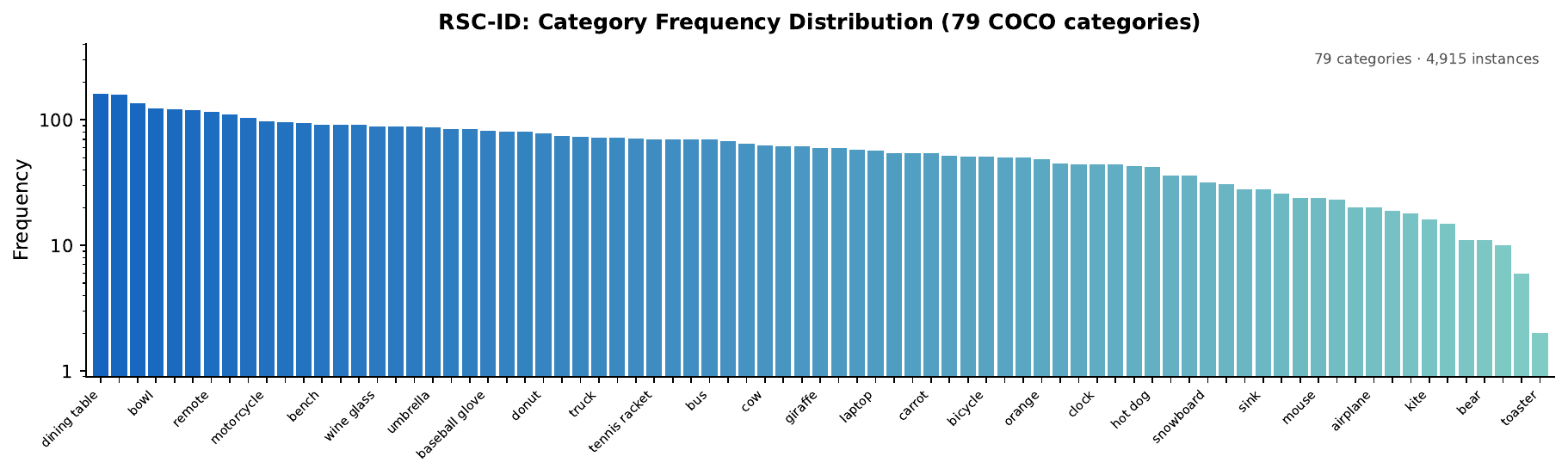}
\caption{\textbf{RSC-ID category frequency distribution.}
Frequency of all 79 COCO categories across 4,038 in-distribution test instances, plotted on a log scale. The distribution is moderately long-tailed, with common household objects dominating.}
\label{fig:category_freq_id}
\end{figure*}

\begin{figure*}[t]
\centering
\includegraphics[width=\linewidth]{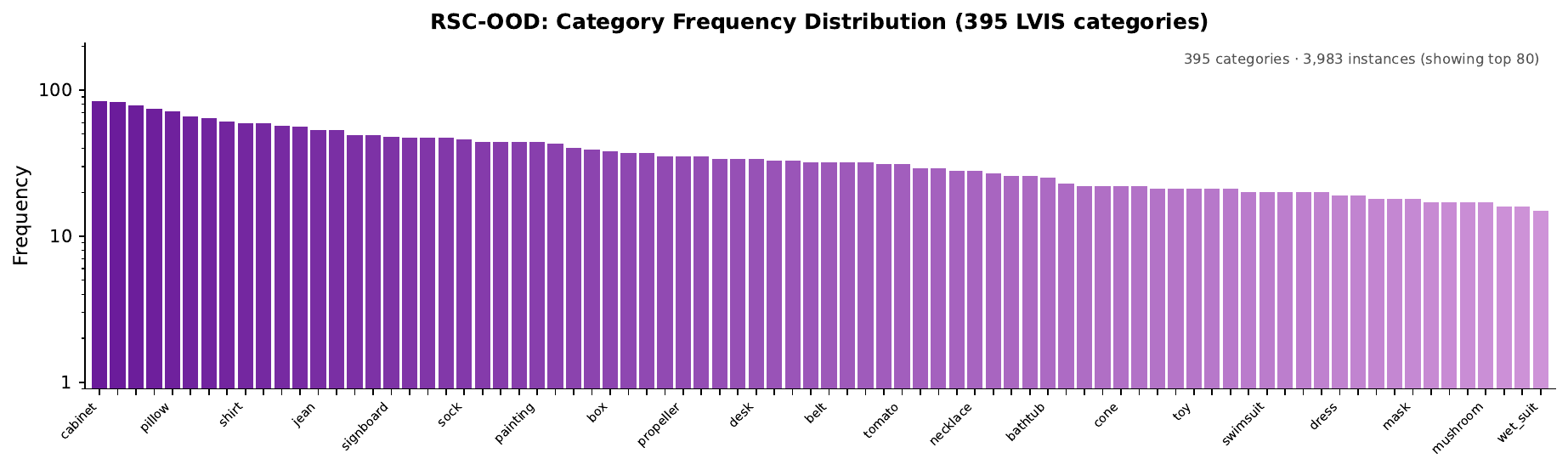}
\caption{\textbf{RSC-OOD category frequency distribution.}
Frequency of the top 80 most common categories among 395 LVIS categories across 3,247 out-of-distribution test instances, plotted on a log scale. The long-tailed distribution reflects LVIS's diverse fine-grained vocabulary.}
\label{fig:category_freq_ood}
\end{figure*}

\paragraph{RSC category distribution.}
Figure~\ref{tab:category_top20} lists the 20 most frequent categories in each test split, and Figures~\ref{fig:category_freq_id} and \ref{fig:category_freq_ood} show the top 80 frequency distributions. RSC-ID covers 79 COCO categories across 4,038 instances, with a moderately long-tailed distribution dominated by common household objects such as \emph{dining table}, \emph{cup}, and \emph{chair}.
RSC-OOD draws from 395 LVIS categories across 3,247 instances, with a substantially longer-tailed distribution reflecting LVIS's fine-grained vocabulary, where top categories include clothing items (\emph{jersey}, \emph{jean}, \emph{shirt}) and household furnishings (\emph{cabinet}, \emph{pillow}, \emph{curtain}). 
Crucially, all OOD categories are disjoint from RSC-ID at both string and synset levels, ensuring that OOD evaluation measures genuine cross-category generalization rather than near-duplicate transfer.

\paragraph{Comparison to natural referential language.}
RSC's linguistic profile aligns with natural referential tendencies identified in prior work. RefCOCOg, the most naturalistically annotated standard benchmark, has an average query length of 8.4 words and a vocabulary of 4,917 unique tokens; RSC extends this tendency toward longer, richer descriptions (52.7 words, 9,086 tokens), consistent with the shift from object-naming to situation-description that characterizes more complex referential contexts. 
The dominance of relational verbs (\emph{distinguish}, \emph{contrast}, \emph{highlight}) and spatial prepositions (\emph{behind}, \emph{beside}, \emph{from}) in RSC queries mirrors patterns documented in human referential communication studies, where disambiguating reference in cluttered scenes naturally recruits relational and contextual cues rather than bare category names~\cite{clark1986referring}. 
While RSC scenarios are LLM-generated, the linguistic structure they exhibit reflects these well-documented tendencies rather than purely synthetic artifacts.

To assess ecological validity, we collected human-authored scenario queries for a subset of 100 RSC instances, asking three annotators to describe the target object without naming its category. Writing a scenario took approximately three minutes per instance, and annotators reported finding the task genuinely challenging: composing a coherent third-person description that provides sufficient disambiguating cues without revealing the target category requires the same kind of intentional reasoning that RSC is designed to evaluate. 
Comparing human-authored and LLM-generated scenarios for the same targets reveals both differences and commonalities. Human scenarios favor the present tense and exhibit more diverse structural patterns, whereas LLM-generated scenarios follow a more consistent template, which typically opens with a role description, followed by distractor contrasts and attribute cues. 
Annotators note that synthetic scenarios are less stylistically varied but more detailed and are factually accurate. 

To verify that both query types convey equivalent referential content, we asked the three annotators to localize the target from the LLM-generated scenario for the same instances they had authored queries for. All annotators successfully identified the correct target from LLM-generated scenarios, confirming that the referential intent is recoverable from either formulation. Example pairs for the same targets are shown below (Table~\ref{tab:human_vs_llm}), illustrating that while surface form differs, the underlying disambiguating content is equivalent. 
Table~\ref{tab:human_vs_llm_eval} shows that model rankings are preserved across both query types and absolute performance levels are comparable, supporting the claim that RSC's LLM-generated scenarios probe the same grounding capability as human-authored ones.
RSC is designed to test whether models can perform this localization task, as long as the scenario accurately identifies the target through relational and contextual cues, whether human- or LLM-authored.

\begin{table}[t]
\centering
\small
\caption{\textbf{Model performance on human-authored vs.\ LLM-generated scenarios} for the same 100 RSC instances. The human-authored scenario query is randomly drawn from one of the three annotators. \method\ performs comparably on human-authored queries, supporting ecological validity and confirming that training does not overfit to LLM-generated patterns.}
\label{tab:human_vs_llm_eval}
\begin{tabular}{lcccc}
\toprule
& \multicolumn{2}{c}{LLM-generated} 
& \multicolumn{2}{c}{Human-authored} \\
\cmidrule(lr){2-3}\cmidrule(lr){4-5}
Method & mIoU & Cat. Acc & mIoU & Cat. Acc \\
\midrule
Qwen2.5-VL  & 33.4 & 31.0 & 31.3 & 32.0 \\
\method & 58.0 & 96.0 & 59.2 & 98.0 \\
\bottomrule
\end{tabular}
\end{table}

\begin{table}[h]
\centering
\small
\caption{\textbf{Human-authored vs.\ LLM-generated 
scenario pairs} for the same targets. Both convey 
equivalent referential content through different 
surface forms.}
\label{tab:human_vs_llm}
\begin{tabular}{p{0.47\linewidth} p{0.47\linewidth}}
\toprule
\textbf{Human-authored} & \textbf{LLM-generated} \\
\midrule
There is a tourist walking down the street who needs to cross at the intersection. You are looking for a green light on a pole, away from the main intersection light. It's closer to the left side of the street and looks higher than the building.
&
As a pedestrian waiting to cross, you want to check if it's safe. You look for the green-lit signal on the rightmost pole, above a blue sign, away from the main intersection lights. Its color, position, and separation from the cluster help you identify it.
\\
\addlinespace[4pt]
A parent is watching from the beach, making sure the kid with dark hair is safe. You notice most of his body is submerged, he's close to an adult, and he's the leftmost in the group. Please identify the kid in the crowd.
&
As a parent watching from the shore, you want to check if your child, who has dark hair and is the leftmost in the group, is safe. They are closest to the water's edge, mostly submerged, and positioned to the left of the shirtless swimmer.
\\
\addlinespace[4pt]
A student is organizing the desk. They are looking for a green-handled item that sits on the desk. There are orange scissors and colorful pens in a mug, but those are not needed. Find the target object to the left of the mug that can be used to cut things.
&
As a teacher organizing supplies, you need the small green-handled item lying on the desk, not the larger orange-handled ones in the mug. Its bright green color, flat position, and separation from the mug make it easy to spot.
\\
\bottomrule
\end{tabular}
\end{table}

\section{RSC Curation Details}
\label{app:rsc_curation}
\begin{figure}[t]
    \centering
    \includegraphics[width=\linewidth]{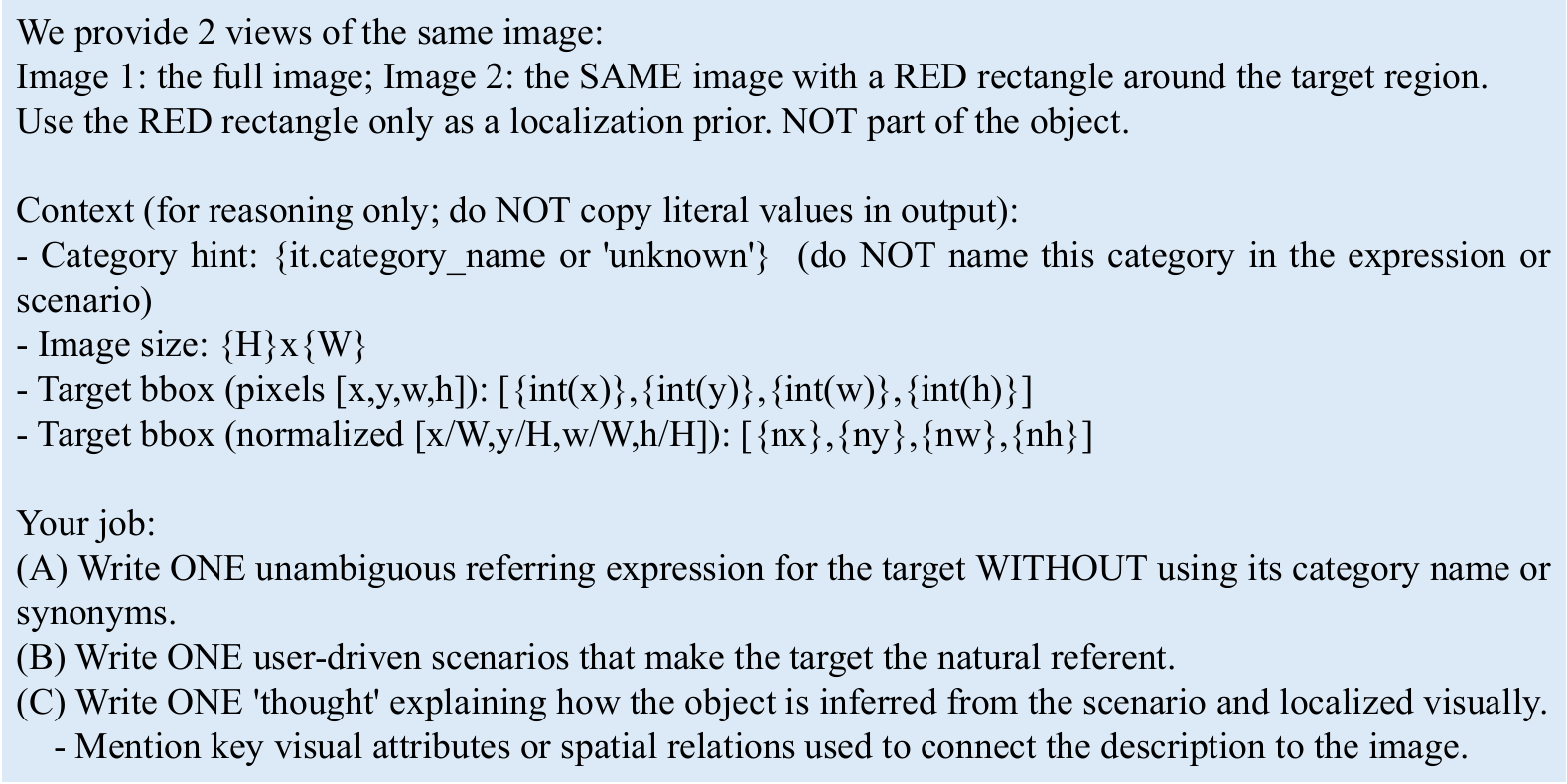}
    \includegraphics[width=\linewidth]{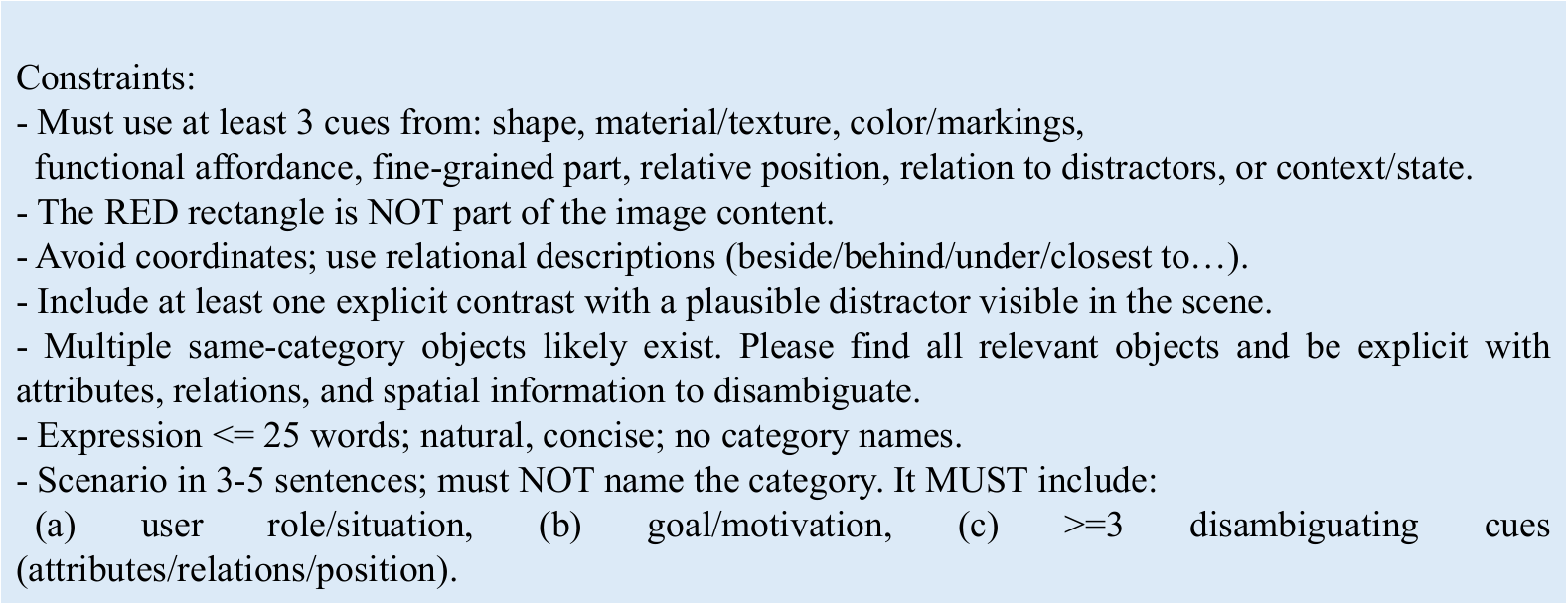}
    \includegraphics[width=\linewidth]{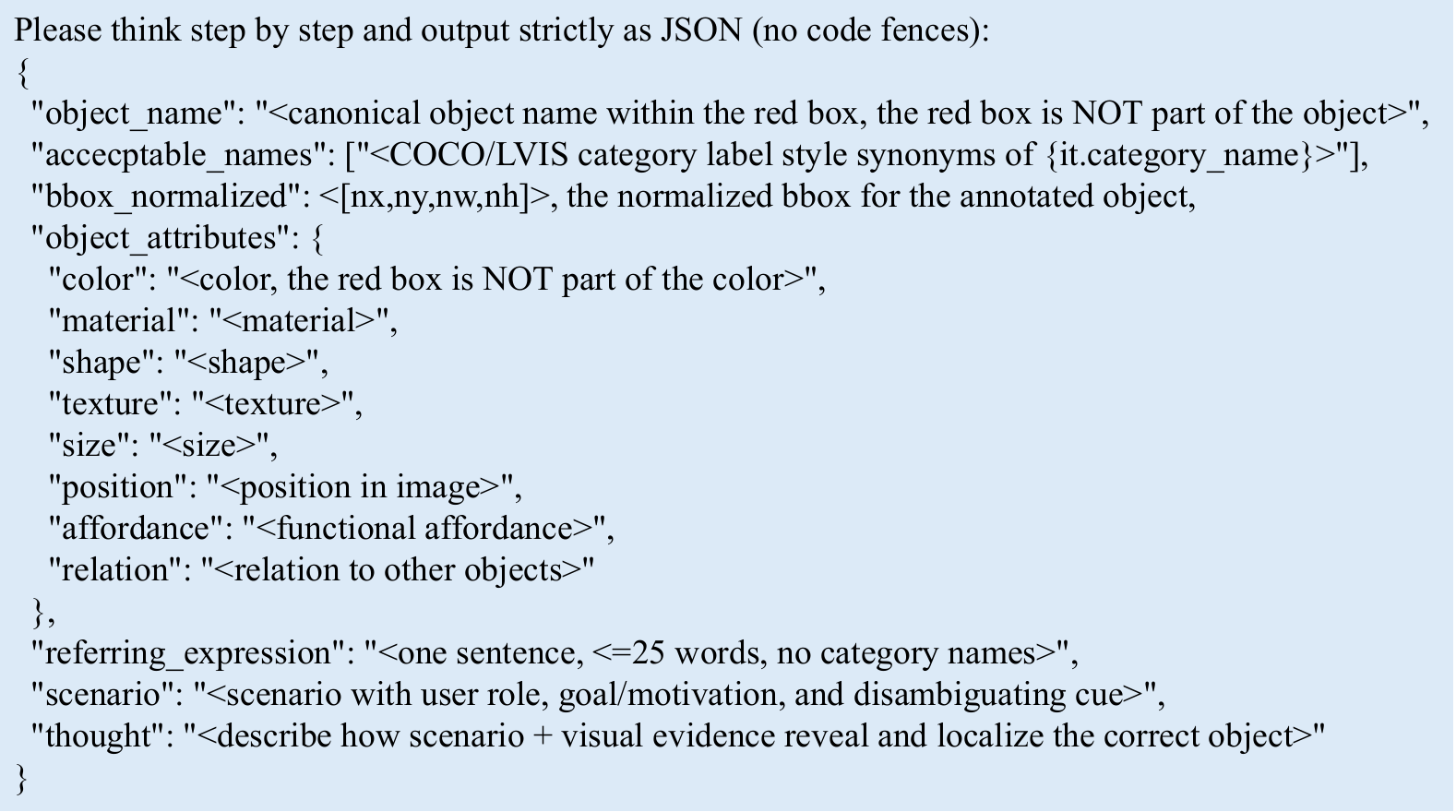}
    \caption{The prompt for Referring Scenario Comprehension curation.}
    \label{fig:rsc_prompt}
\end{figure}

\paragraph{Annotation prompts.}
Fig.~\ref{fig:rsc_prompt} demonstrates the prompt used for RSC annotation generation. For each target, the annotator LLM is shown two views of the \emph{same} image—the full frame and the frame with a red rectangle prior \(R(b^\ast)\) over the target region (the rectangle is \emph{not} part of the object). The prompt supplies context for reasoning only (category hint, image size \(H{\times}W\), pixel box \([x,y,w,h]\), and its normalized form \([x/W,y/H,w/H,h/H]\)) and then instructs the model to (i) write one \emph{category-free} referring expression (\(\le\)25 words), (ii) write one user-driven \emph{scenario} (3–5 sentences specifying role/goal and \(\ge\)3 disambiguating cues from attributes/relations/position/affordance), and (iii) produce a brief \emph{thought} explaining how the scenario maps to visual evidence. Outputs must be \emph{JSON-only} with the fixed schema: \texttt{object\_name}, \texttt{acceptable\_names} (alias list for internal validation), \texttt{bbox\_normalized} \([\tfrac{x}{W},\tfrac{y}{H},\tfrac{w}{W},\tfrac{h}{H}]\), \texttt{object\_attributes} \{\texttt{color}, \texttt{material}, \texttt{shape}, \texttt{texture}, \texttt{size}, \texttt{position}, \texttt{affordance}, \texttt{relation}\}, \texttt{referring\_expression}, \texttt{scenario}, and \texttt{thought}. The template enforces JSON validity, forbids category leakage in the expression/scenario, requires at least one explicit contrast with plausible distractors, and discourages coordinates in prose in favor of relational cues.

\paragraph{Filtering and quality judge.}
We first provide the predicted box and the GT bounding boxes from Phase 2 to compute the IoU as the first step of filtering. Then we provide the image, category label, and the generated acceptable name list to GPT-4o~\cite{hurst2024gpt} for the second step of filtering.
To reduce the risk of annotation leakage from the localization prior, we not only explicitly instruct the generator to avoid coordinate-based descriptions in favor of relational cues (e.g., \emph{beside}, \emph{behind}, \emph{closest to}), but also apply a leakage detection to reject any scenario that references the rectangle or uses spatial coordinates verbatim. 

\begin{figure}[t]
\centering
\includegraphics[width=\linewidth]{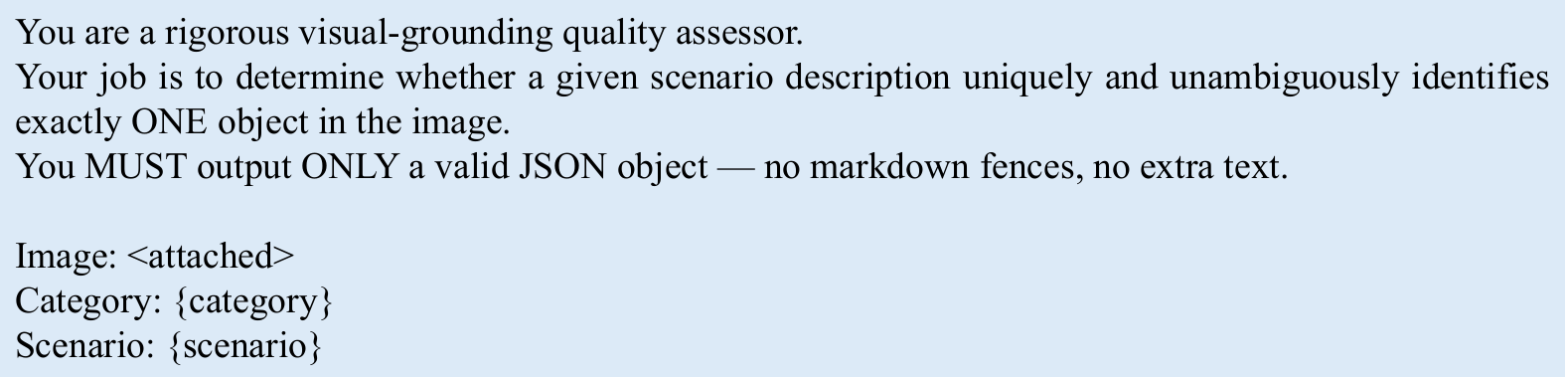}
\includegraphics[width=\linewidth]{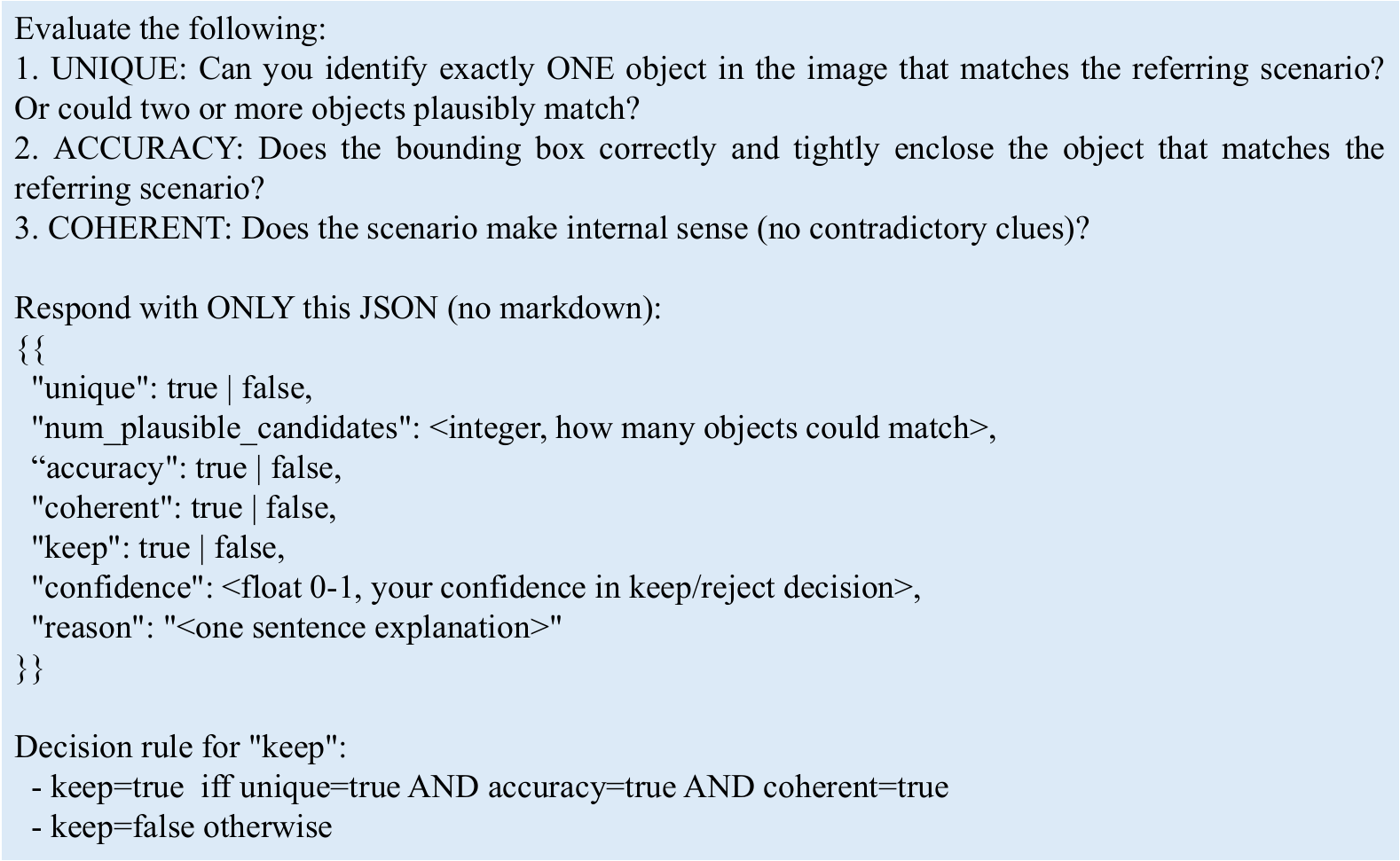}
\caption{\textbf{Quality judge prompt.} The judge assesses three criteria: uniqueness, bounding box accuracy, and scenario coherence. It applies a strict conjunction rule: an instance is kept only when all three criteria are satisfied.}
\label{fig:judge_prompt}
\end{figure}

Figure~\ref{fig:judge_prompt} shows the prompt used by the quality judge in Phase~3. The system prompt instructs the judge to act as a rigorous visual-grounding quality assessor, determining whether the scenario uniquely and unambiguously identifies exactly one object in the image. The evaluation prompt asks the judge to assess three criteria: \textsc{Unique} (whether exactly one object in the image plausibly matches the scenario), \textsc{Accuracy} (whether the bounding box correctly and tightly encloses the matching object), and \textsc{Coherent} (whether the scenario is internally consistent with no contradictory cues). The judge returns a structured JSON response including per-criterion verdicts, a plausible candidate count, a confidence score, and a one-sentence reason. An instance is kept if and only if all three criteria are satisfied (\textsc{Unique} $\wedge$ \textsc{Accuracy} $\wedge$ \textsc{Coherent}); borderline cases with low confidence enter the judge refinement loop described in Section~\ref{sec:rsc}.

\paragraph{Human audit.}
Human verification follows the same three-criterion protocol as the quality judge: annotators assess \textsc{Unique}, \textsc{Accuracy}, and \textsc{Coherent} for each instance, with a keep decision requiring all three to be satisfied. Three computer science PhD students conducted the audit independently, without knowledge of each other's decisions. To reduce cultural and linguistic bias, annotators were recruited from three different countries (two male, one female). 
Each annotator reviewed a stratified random sample of 100 instances per audit round. Majority-vote accuracy was 95.6\% overall (per-rater: 94\%/96\%/97\%) with great inter-annotator agreement (Fleiss’ $\kappa{=}\,$\textit{0.94}).

\section{Data Examples}
\label{sec:data_examples_supp}
\begin{figure}[t]
    \centering
    \includegraphics[width=\linewidth]{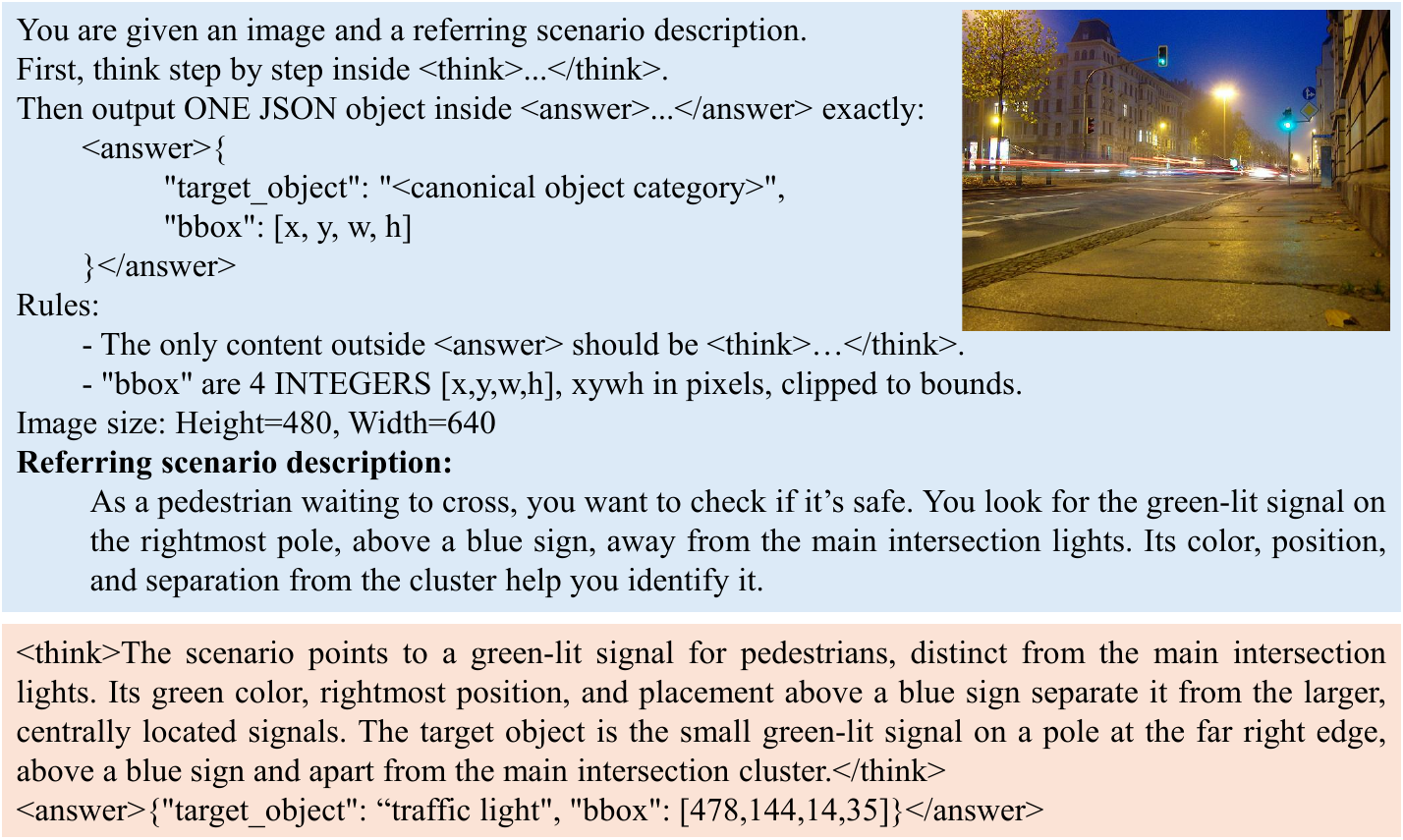}
    \includegraphics[width=\linewidth]{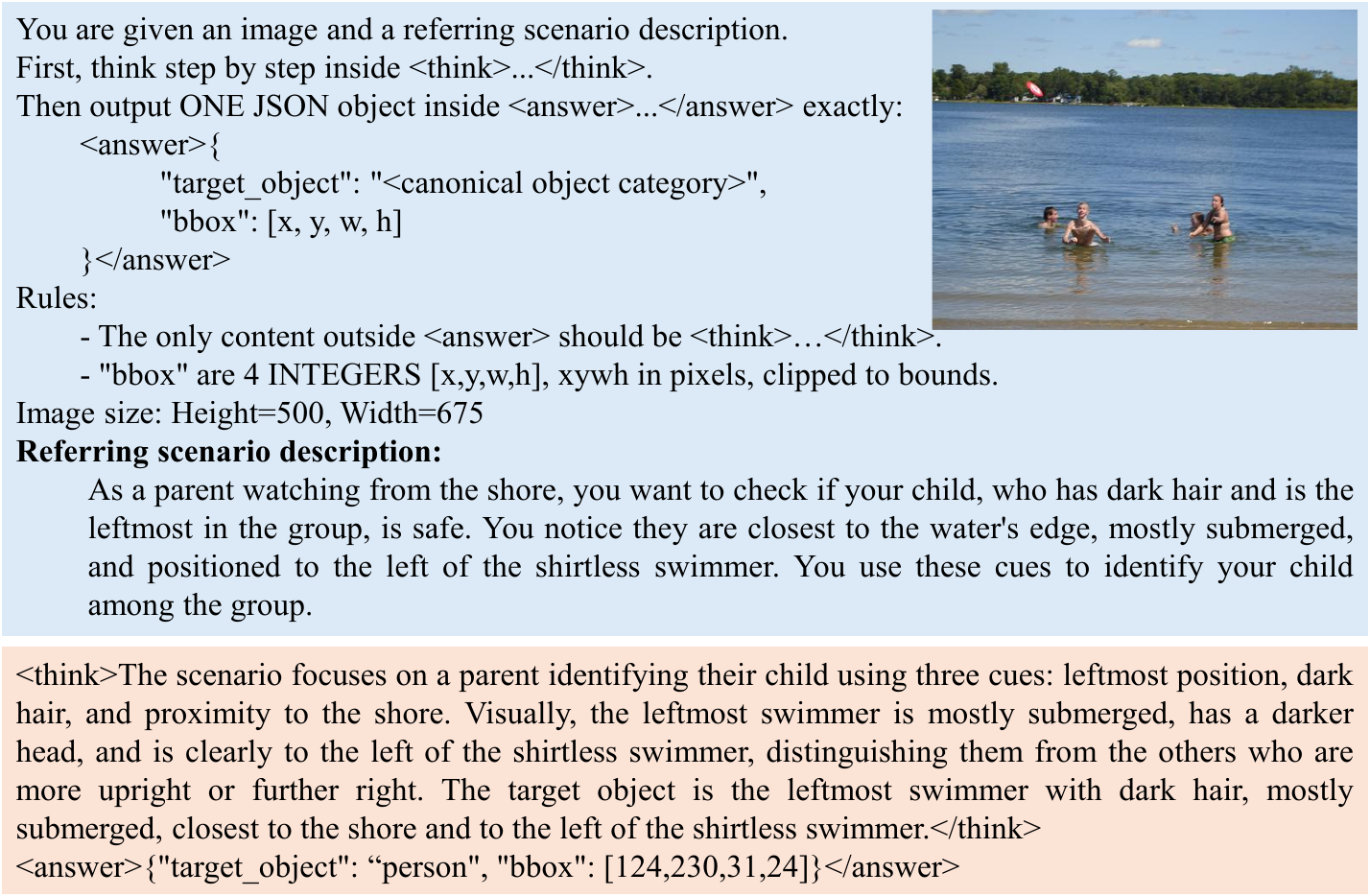}
    \includegraphics[width=\linewidth]{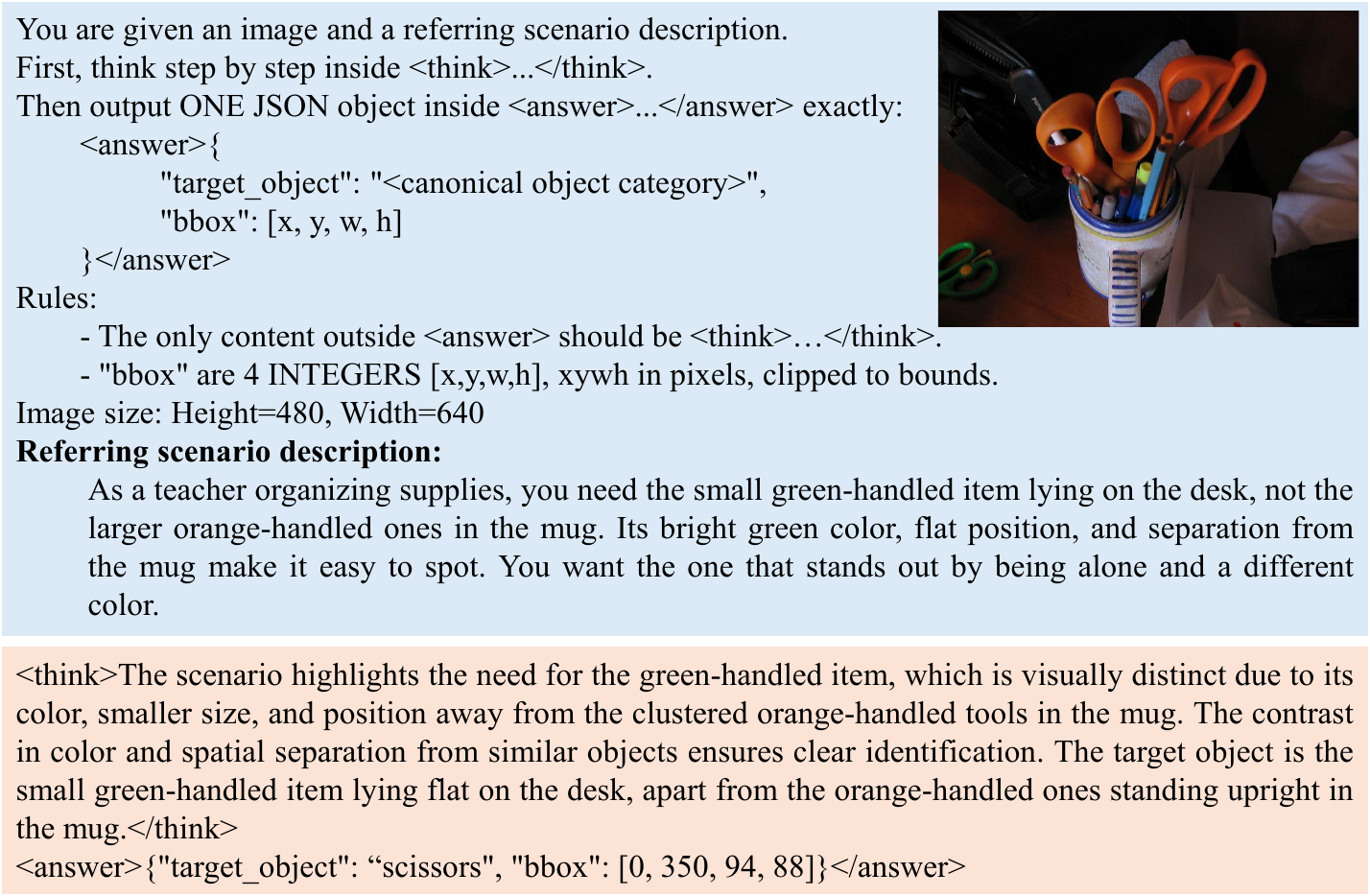}
    \caption{RSC training examples.}
    \label{fig:rsc_reasoning_supp}
\end{figure}

\begin{figure}[t]
    \centering
    \includegraphics[width=\linewidth]{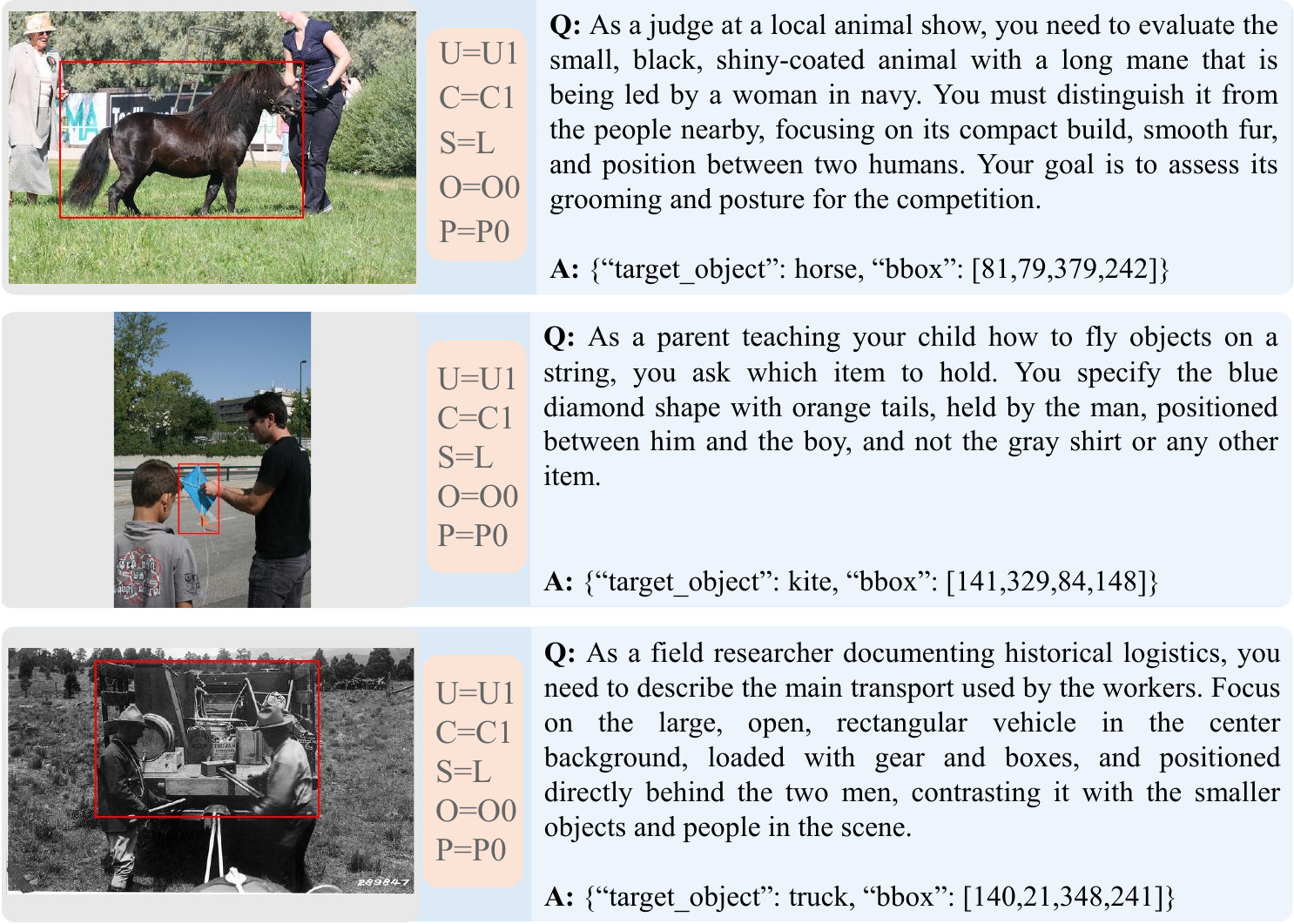}
    \caption{RSC examples with difficulty level "easy".}
    \label{fig:rsc_examples1_supp}
\end{figure}

\begin{figure}[t]
    \centering
    \includegraphics[width=\linewidth]{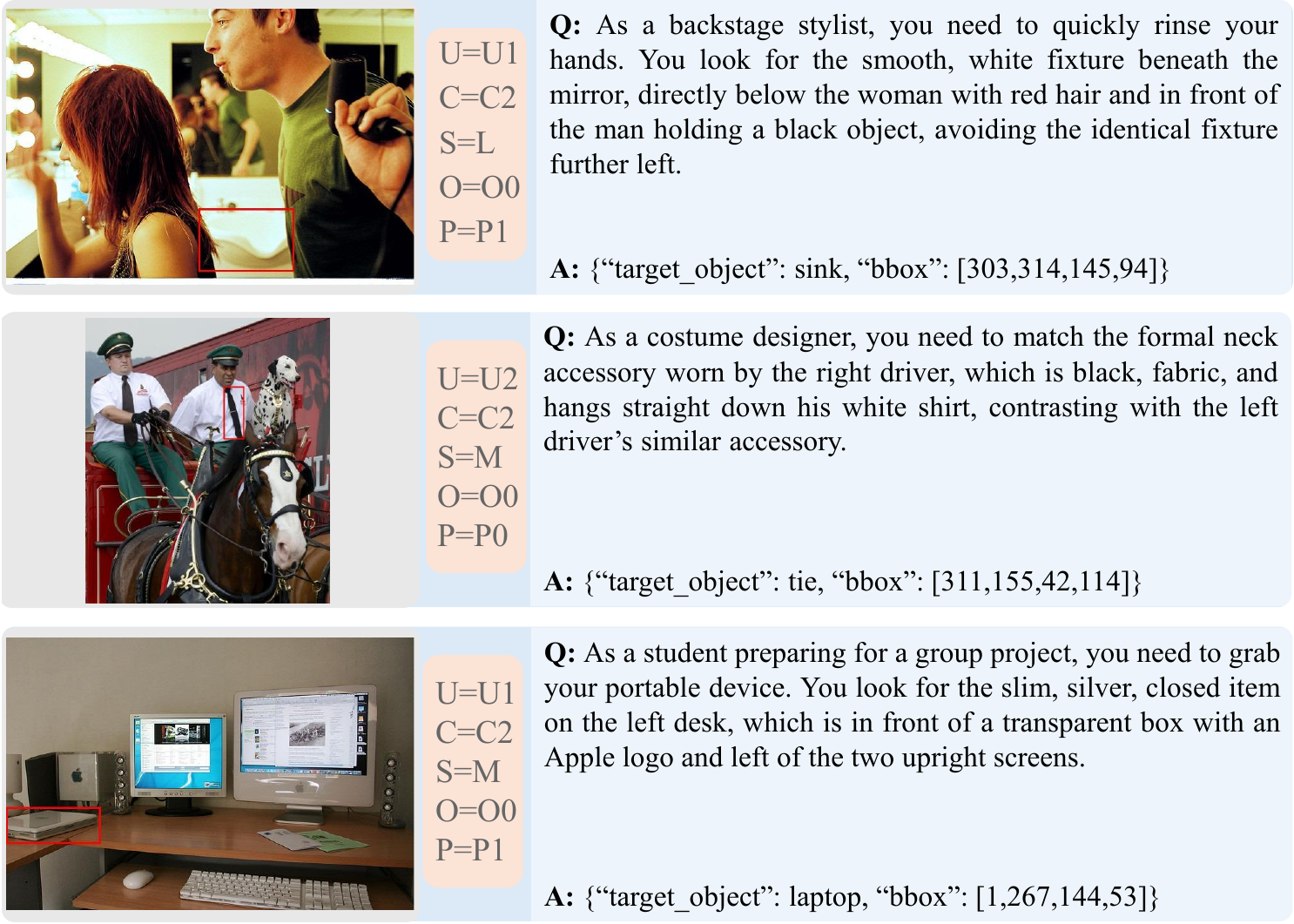}
    \caption{RSC examples with difficulty level "medium".}
    \label{fig:rsc_examples2_supp}
\end{figure}

\begin{figure}[t]
    \centering
    \includegraphics[width=\linewidth]{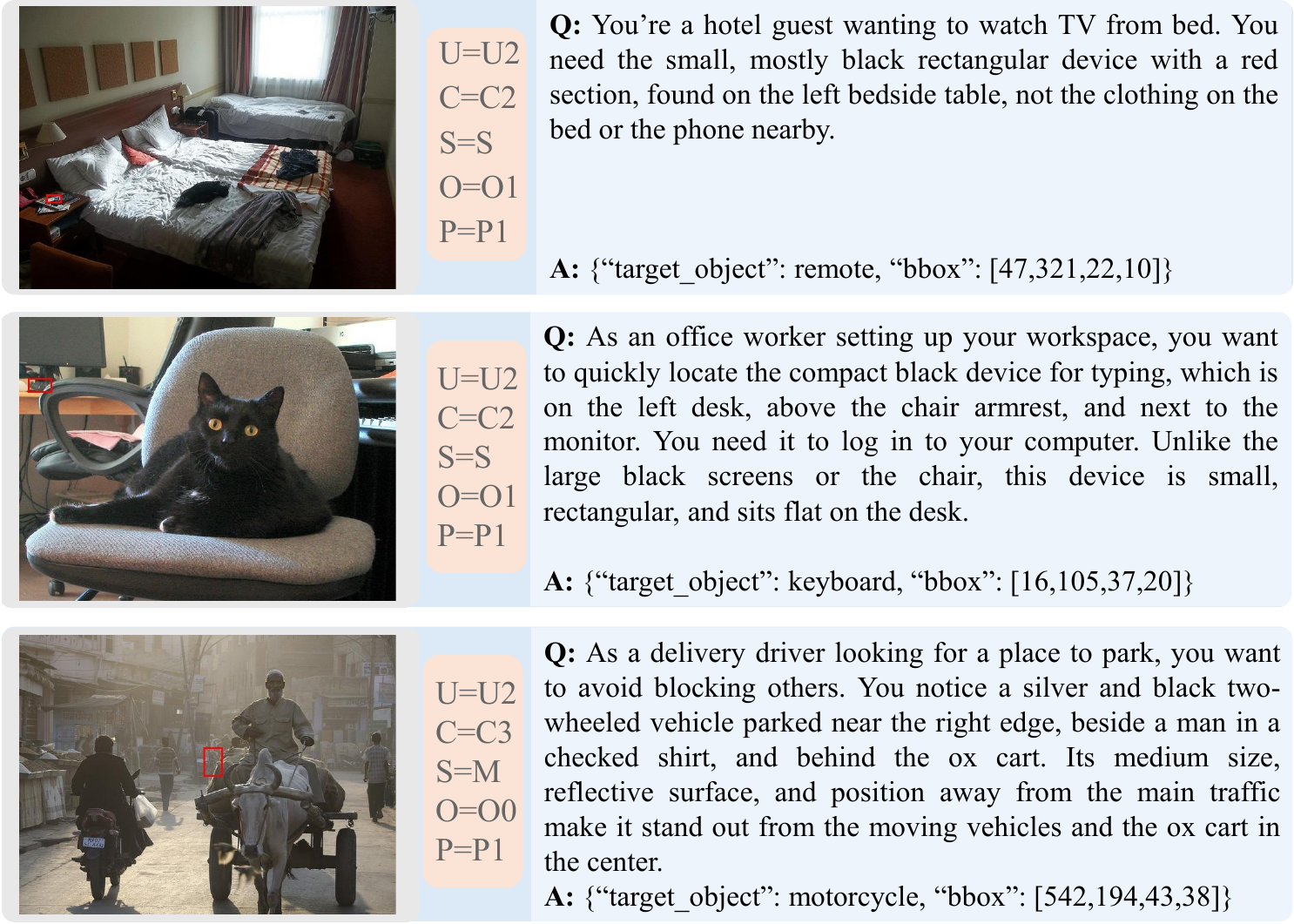}
    \caption{RSC examples with difficulty level "hard".}
    \label{fig:rsc_examples3_supp}
\end{figure}

We include additional data examples to illustrate the user-driven scenarios, reasoning traces, and tag structure in RSC. 
Figure~\ref{fig:rsc_reasoning_supp} samples from the training split and, for each instance, shows the natural image, the user-driven scenario and the reasoning process.
From the held-out test split, Figures~\ref{fig:rsc_examples1_supp}-~\ref{fig:rsc_examples2_supp}-\ref{fig:rsc_examples3_supp} present \emph{easy}, \emph{medium}, and \emph{hard} examples, respectively, reflecting increasing ambiguity (U2), scene clutter (C2/C3), occlusion/overlap (O1/O2), and off-center placement (P1). 
All examples satisfy the curation gates (alias consistency and IoU threshold; see main text) and omit category names to preserve the scenario.

\section{\method: Implementation Details}
\label{app:method}

This appendix gives a complete, symbol-level description of 
\method. All equations correspond one-to-one with our 
implementation; concrete hyperparameter values appear in 
Section~\ref{sec:exp_setup}.

\subsection{Thought-Primed SFT (TP-SFT)}

\paragraph{Objective.}
Given image $x$, scenario $s$, and target text $\mathbf{y}$ 
(the concatenation of \texttt{<think>} and \texttt{<answer>} 
spans), we minimize:
\begin{equation}
\small
\mathcal{L}_{\mathrm{SFT}}(\theta)
= -\mathbb{E}_{(x,s,\mathbf{y})\sim\mathcal{D}_{\mathrm{sft}}}
\left[
  \sum_{t=1}^{|\mathbf{y}|}
  \log p_\theta\!\left(y_t \mid x,s,\mathbf{y}_{<t}\right)
\right].
\label{eq:sft}
\end{equation}

\paragraph{Output schema.}
The instruction requires a single JSON inside 
\texttt{<answer>}\dots\texttt{</answer>} with two keys: 
\texttt{target\_object} (canonical category string) and 
\texttt{bbox} (four pixel integers in xywh, clipped to 
image bounds). Several key aliases are accepted at scoring 
time for robustness.

\paragraph{Curriculum.}
TP-SFT draws from the easy RSC slice (instances with $D_i$ 
below the easy-percentile threshold $\delta_{\mathrm{easy}}$) 
to stabilize schema and trace learning before RL. The trained 
checkpoint is saved as the reference policy $\pi_{\mathrm{ref}}$ 
that anchors KL regularization in Stage~2.

\subsection{Incentive-Curriculum GRPO (IC-GRPO)}

\paragraph{GRPO objective.}
For each item $(x_i, s_i)$ we sample $K$ completions 
$\{c_{i,k}\}_{k=1}^K \sim \pi_\theta(\cdot \mid x_i,s_i)$ 
and compute group-relative advantages:
\begin{equation}
\small
A_{i,k} = r_{i,k} 
  - \frac{1}{K}\sum_{k'=1}^{K} r_{i,k'}.
\label{eq:advantage}
\end{equation}
The KL-regularized objective is:
\begin{align}
\small
\mathcal{J}(\theta)
=& \mathbb{E}_i\!\left[
  \frac{1}{K}\sum_{k=1}^{K}
  A_{i,k}\log\pi_\theta\!\left(c_{i,k}\mid x_i,s_i\right)
\right] \nonumber \\
&- \beta\,\mathrm{KL}\!\left(\pi_\theta \,\|\, \pi_{\mathrm{ref}}\right).
\label{eq:grpo}
\end{align}
The KL coefficient $\beta$ is adapted online by the 
\texttt{AdaptiveKLScheduler}: if the observed KL exceeds 
the target band $[\kappa_{\mathrm{tgt}} - \kappa_{\mathrm{tol}},\, 
\kappa_{\mathrm{tgt}} + \kappa_{\mathrm{tol}}]$, $\beta$ is 
multiplied by $\mu_{\uparrow}$; if it falls below, by 
$\mu_{\downarrow}$; and $\beta$ is clipped to 
$[\beta_{\min}, \beta_{\max}]$.

\paragraph{Box normalization.}
A predicted 4-vector $\widehat{b} = [x,y,a,b']$ is mapped 
to a valid box $\widetilde{b} \in \mathbb{Z}_{\ge0}^4$ by 
the following priority logic:
\begin{enumerate}
  \item \textbf{Prefer xywh:} if $(x,y)$ lie inside the 
  image and $(a,b')$ are valid width/height that fit within 
  bounds, interpret as $(x,y,w,h)$.
  \item \textbf{Try xyxy:} else if $a > x$ and $b' > y$ 
  and all values are within image bounds, convert 
  $(x_1,y_1,x_2,y_2)$ to xywh.
  \item \textbf{Clamp:} otherwise clamp aggressively to 
  image bounds with minimum $1$-px side length.
\end{enumerate}
The out-of-bounds indicator is 
$\mathbb{1}_{\mathrm{OOB}} = \mathbf{1}\{\widehat{b} \neq \widetilde{b}\}$.
All IoU values are computed in xyxy after conversion.

\paragraph{Geometry reward.}
Let $\sigma(z) = (1+e^{-z})^{-1}$ be the logistic function, 
$c(\cdot)$ the box center, and 
$\mathrm{diag} = \sqrt{W^2+H^2}$ the image diagonal. 
Define the normalized center distance 
$d = \|c(\widetilde{b}) - c(b^\ast)\|_2 / \mathrm{diag}$.
The geometry reward is:
\begin{align}
\small
r_{\mathrm{iou}}
&= \min\!\Big\{1,\;
  \underbrace{\mathrm{IoU}(\widetilde{b}, b^\ast)}_{\text{base}}
  + \alpha_1\,\sigma\!\left(\tfrac{\mathrm{IoU}(\widetilde{b},b^\ast)-\tau_1}{\kappa}\right)
  \nonumber \\
&\quad\quad
  + \alpha_2\,\sigma\!\left(\tfrac{\mathrm{IoU}(\widetilde{b},b^\ast)-\tau_2}{\kappa}\right)
  + \alpha_c\,\exp\!\left(-\tfrac{d^2}{2\sigma_c^2}\right)
\Big\} \nonumber \\
&\quad\quad
  - \alpha_{\mathrm{oob}}\,\mathbb{1}_{\mathrm{OOB}},
\label{eq:riou}
\end{align}
where $\tau_1 < \tau_2$ are the two IoU operating points, 
$\kappa$ controls logistic width, $\alpha_1, \alpha_2$ 
are the corresponding bonus magnitudes, $\alpha_c$ and 
$\sigma_c$ govern the center-consistency term, and 
$\alpha_{\mathrm{oob}}$ is the out-of-bounds penalty.

\paragraph{Category reward.}
Let $\mathcal{A}^{\mathrm{can}}$ be the canonical name set 
and $\mathcal{A}^{\mathrm{norm}}$ the normalized alias set. 
Define an IoU gate:
\begin{equation}
\small
g_{\mathrm{iou}} =
\begin{cases}
1, & \mathrm{IoU}(\widetilde{b},b^\ast) \ge \tau_g, \\
g, & \mathrm{IoU}(\widetilde{b},b^\ast) < \tau_g,
\end{cases}
\end{equation}
where $\tau_g$ is the gate threshold and $g < 1$ is the 
halving factor. With token-level Jaccard similarity 
$\mathrm{Jac}(\cdot,\cdot)$ computed after lowercasing 
and light stemming:
\begin{equation}
\small
r_{\mathrm{cat}}
= g_{\mathrm{iou}} \times
\begin{cases}
1, & \widehat{y} \in \mathcal{A}^{\mathrm{can}}, \\[3pt]
\eta, & \widehat{y} \in \mathcal{A}^{\mathrm{norm}}, \\[3pt]
\rho_l + \rho_s\,\max\limits_{a\in\mathcal{A}^{\mathrm{norm}}}
  \mathrm{Jac}(\widehat{y},a), & \text{otherwise,}
\end{cases}
\label{eq:rcat}
\end{equation}
where $\eta$ is the alias credit, and $\rho_l, \rho_s$ 
map Jaccard $\in [0,1]$ to the soft-overlap range 
$[\rho_l,\, \rho_l + \rho_s]$.

\paragraph{Format and structure rewards.}
Let $\delta_{\mathrm{tag}} \in \{0,1\}$ indicate the 
presence of answer tags, $\delta_{\mathrm{json}} \in \{0,1\}$ 
whether a parseable JSON is found inside, and 
$\delta_{\mathrm{keys}} \in \{0,1\}$ whether both a bbox 
key and a class key are present:
\begin{align}
\small
r_{\mathrm{fmt}} &=
\begin{cases}
+1, & \delta_{\mathrm{tag}} = 1 \;\text{and}\; \delta_{\mathrm{json}} = 1, \\
-1, & \text{otherwise,}
\end{cases}
\label{eq:rfmt} \\[4pt]
r_{\mathrm{struct}} &=
\max\!\left\{
  \gamma_{\mathrm{tag}}\,\delta_{\mathrm{tag}}
  + \gamma_{\mathrm{key}}\,\delta_{\mathrm{keys}},\;
  \gamma_{\min}
\right\},
\label{eq:rstruct}
\end{align}
where $\gamma_{\mathrm{tag}}$, $\gamma_{\mathrm{key}}$, and 
$\gamma_{\min}$ are the tag, key, and floor coefficients.

\paragraph{Total reward and weight annealing.}
The scalar reward is:
\begin{equation}
\small
r = w_{\mathrm{iou}}\,r_{\mathrm{iou}}
  + w_{\mathrm{cat}}\,r_{\mathrm{cat}}
  + w_{\mathrm{fmt}}\,r_{\mathrm{fmt}}
  + w_{\mathrm{struct}}\,r_{\mathrm{struct}}.
\label{eq:rtotal}
\end{equation}
Weights are linearly annealed from start to late values 
over the first $p_{\mathrm{anneal}}$ fraction of training:
\begin{align}
\small
w_h(s) = w_h^{\mathrm{start}}
  + p(s)\,\bigl(w_h^{\mathrm{late}} - w_h^{\mathrm{start}}\bigr), \\
\nonumber
p(s) = \min\!\left\{1,\,\frac{s}{p_{\mathrm{anneal}} \cdot S}\right\},
\label{eq:anneal}
\end{align}
where $s$ is the global step and $S$ the total steps.

\paragraph{Tag-aware curriculum.}
IC-GRPO samples RSC using difficulty scores $D_i$ and 
tag-based marginals. Let 
$\Pi^{(m)} = (\pi^{(m)}_{\mathrm{easy}},\, 
\pi^{(m)}_{\mathrm{med}},\, \pi^{(m)}_{\mathrm{hard}})$ 
denote the easy/medium/hard sampling mixture for RL stage 
$m$. The mixture shifts between stages to progressively 
expose the policy to harder U2, C3, O2, and P1 instances, 
addressing reward sparsity by first establishing reliable 
IoU signals on easier cases.

\paragraph{Prompt-template ensemble (PTE-8).}
At each training step we uniformly sample one of $T$ 
prompt paraphrases. All templates enforce the same output 
schema; rewards are logged per template for analysis but 
supervision is identical, improving robustness to surface 
query variation without changing the learning objective.

\section{Tag-level Analysis}
\label{app:tag}
Table~\ref{tab:tag_analysis_ood} breaks down performance by difficulty tag on RSC-OOD. For mIoU, GRPO consistently improves over SFT across all tags, with the largest relative gains on small instances (S: 2.56$\to$17.18) and high-overlap instances (O2: 29.40$\to$43.74), confirming that the difficulty-aware curriculum meaningfully improves localization on the hardest cases. Large objects (L) achieve the highest absolute mIoU (60.67), as expected from their visual salience. For category accuracy, a notable pattern emerges: SFT drops below the baseline across nearly all tags, indicating that schema alignment reduces OOD category generalisation. GRPO recovers and surpasses the baseline in most cases, demonstrating that RL restores and extends semantic understanding on unseen categories beyond what SFT alone provides. High-overlap instances (O2) remain hard for category naming across all methods, suggesting that visual congestion impairs both localisation and semantic inference simultaneously.

\begin{table}[t]
\centering
\small
\caption{\textbf{Tag-level analysis on RSC-OOD.}
mIoU and category accuracy (Cat Acc) broken down 
by difficulty tag for the baseline (Qwen2.5-VL), 
SFT, and GRPO.}
\label{tab:tag_analysis_ood}
\setlength{\tabcolsep}{4pt}
\resizebox{\linewidth}{!}{
\begin{tabular}{lcccccc}
\toprule
& \multicolumn{3}{c}{mIoU (\%)} 
& \multicolumn{3}{c}{Cat Acc (\%)} \\
\cmidrule(lr){2-4}\cmidrule(lr){5-7}
Tag & Baseline & SFT & GRPO 
    & Baseline & SFT & GRPO \\
\midrule
U1 & 27.24 & 41.71 & 46.78 & 26.24 & 13.57 & 22.20 \\
U2 & 16.98 & 26.44 & 31.67 & 16.52 & 11.71 & 20.28 \\
\midrule
C1 & 22.85 & 36.06 & 39.54 & 12.32 & 10.53 & 19.26 \\
C2 & 21.30 & 33.00 & 39.29 & 19.93 & 12.22 & 21.33 \\
C3 & 20.87 & 31.44 & 36.40 & 27.82 & 14.31 & 22.18 \\
\midrule
S  &  2.56 & 14.85 & 17.18 & 14.85 & 12.41 & 21.87 \\
M  & 16.50 & 33.49 & 38.78 & 20.56 & 12.60 & 21.00 \\
L  & 49.59 & 52.64 & 60.67 & 27.67 & 12.57 & 20.51 \\
\midrule
O0 & 18.03 & 30.11 & 34.69 & 20.30 & 14.81 & 22.77 \\
O1 & 21.18 & 33.53 & 39.29 & 19.80 & 12.08 & 21.83 \\
O2 & 29.40 & 38.73 & 43.74 & 24.15 &  8.91 & 16.21 \\
\midrule
P0 & 16.38 & 33.01 & 37.36 & 21.34 & 13.96 & 22.38 \\
P1 & 26.77 & 33.39 & 39.38 & 20.29 & 11.08 & 19.85 \\
\bottomrule
\end{tabular}}
\end{table}

\section{Qualitative Examples}
\label{sec:qualitative_examples_supp}

\begin{figure*}[t]
    \centering
    \includegraphics[width=0.48\linewidth]{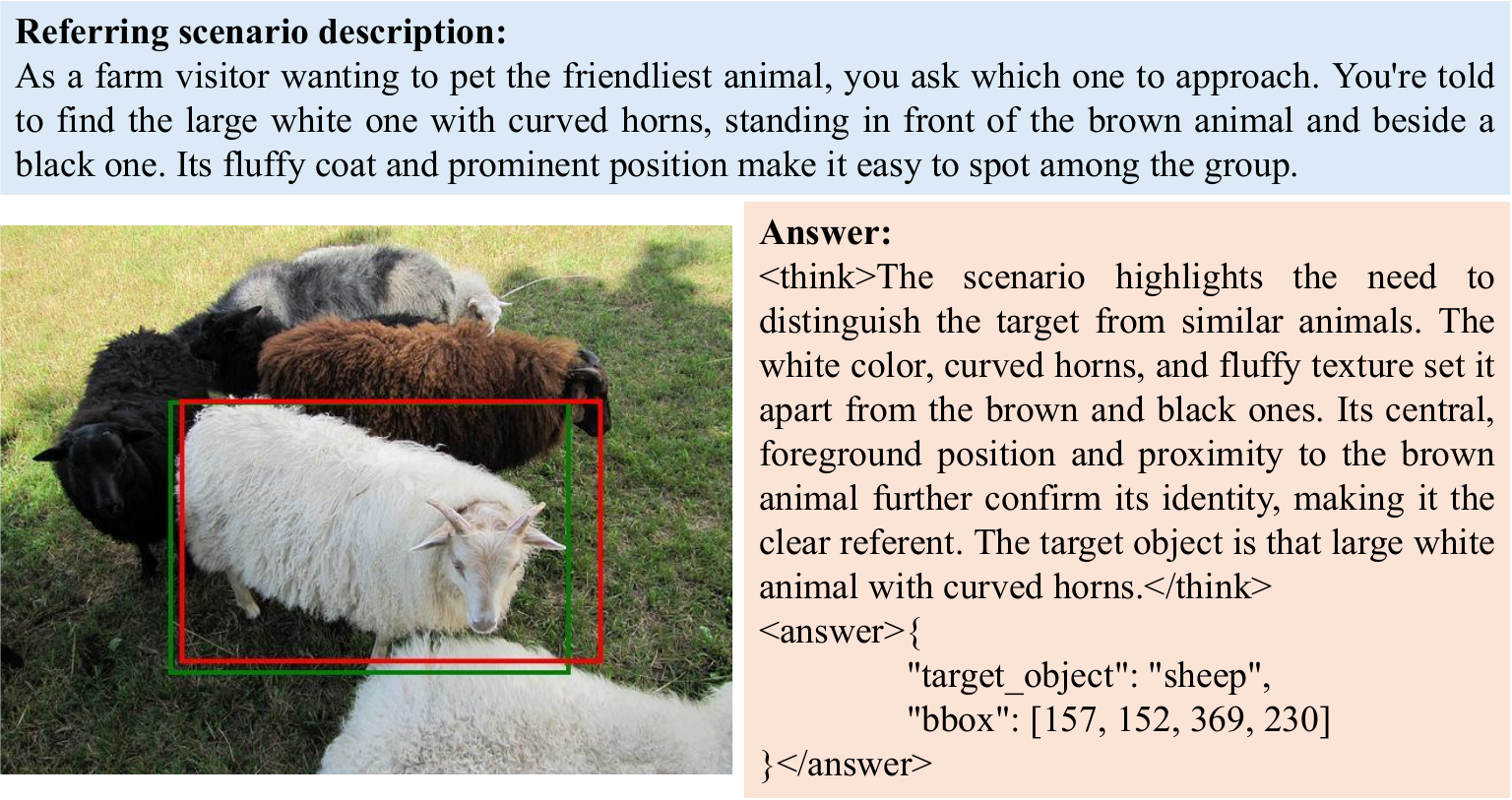}
    \includegraphics[width=0.48\linewidth]{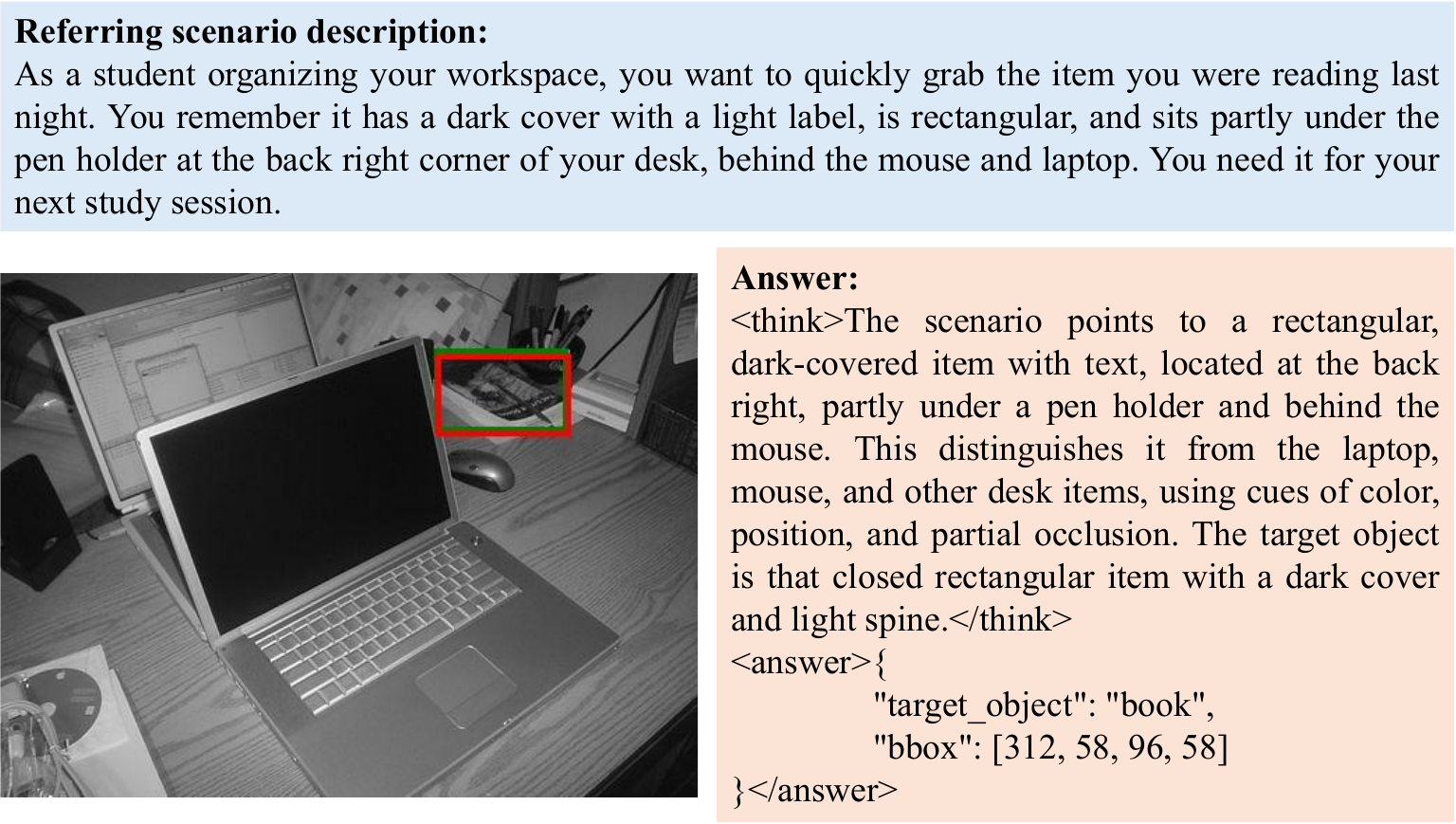}
    \includegraphics[width=0.48\linewidth]{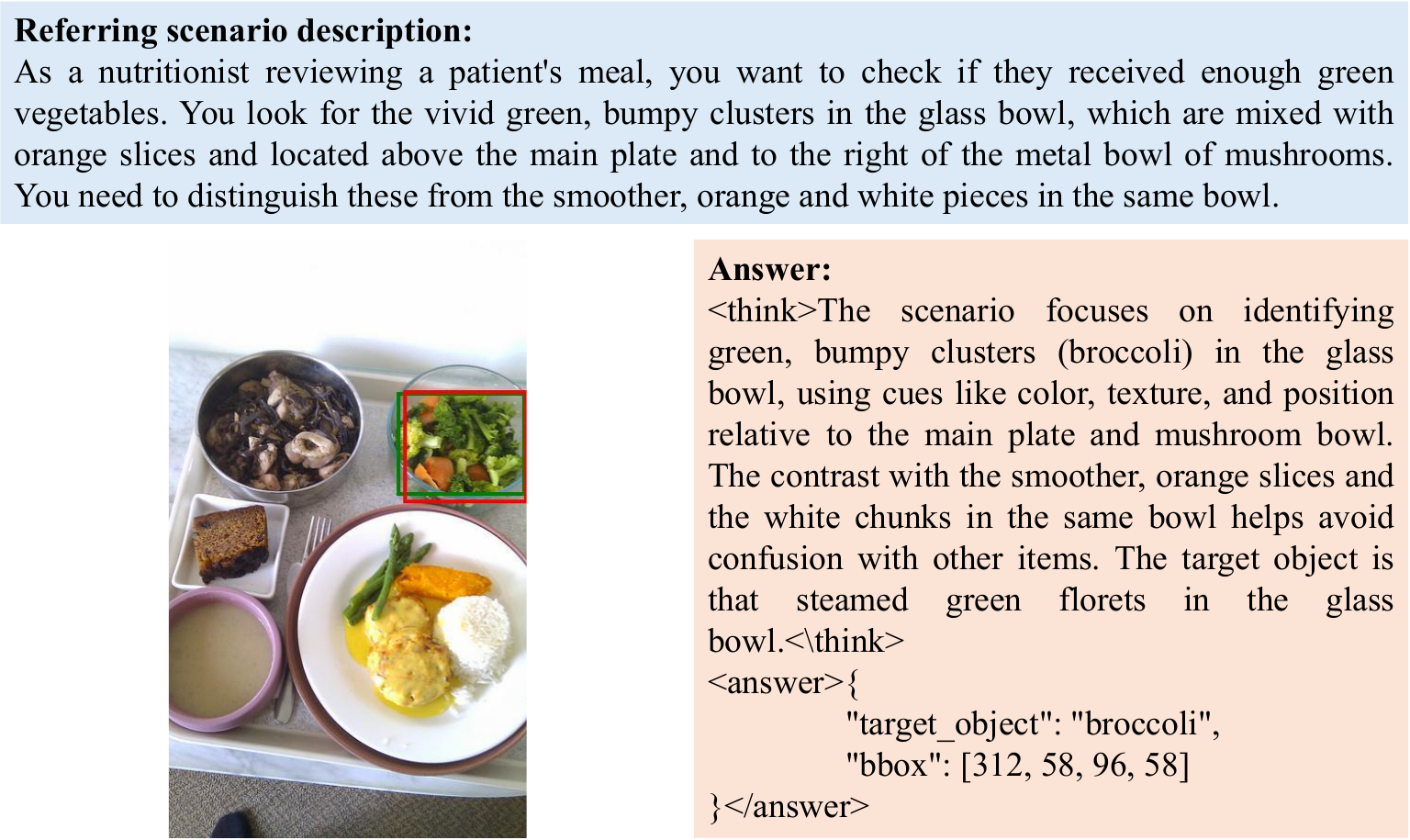}
    \includegraphics[width=0.48\linewidth]{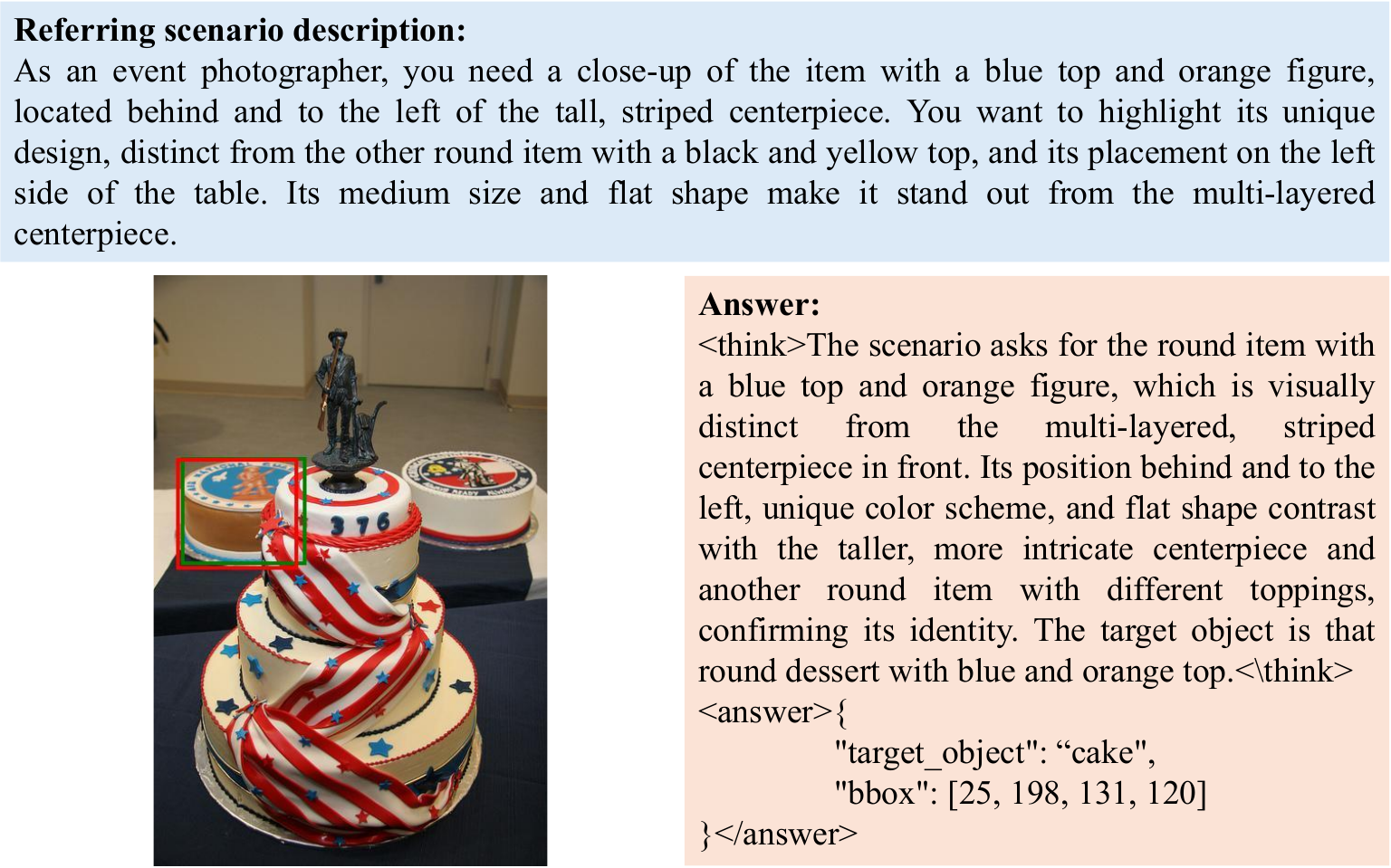}
    \includegraphics[width=0.48\linewidth]{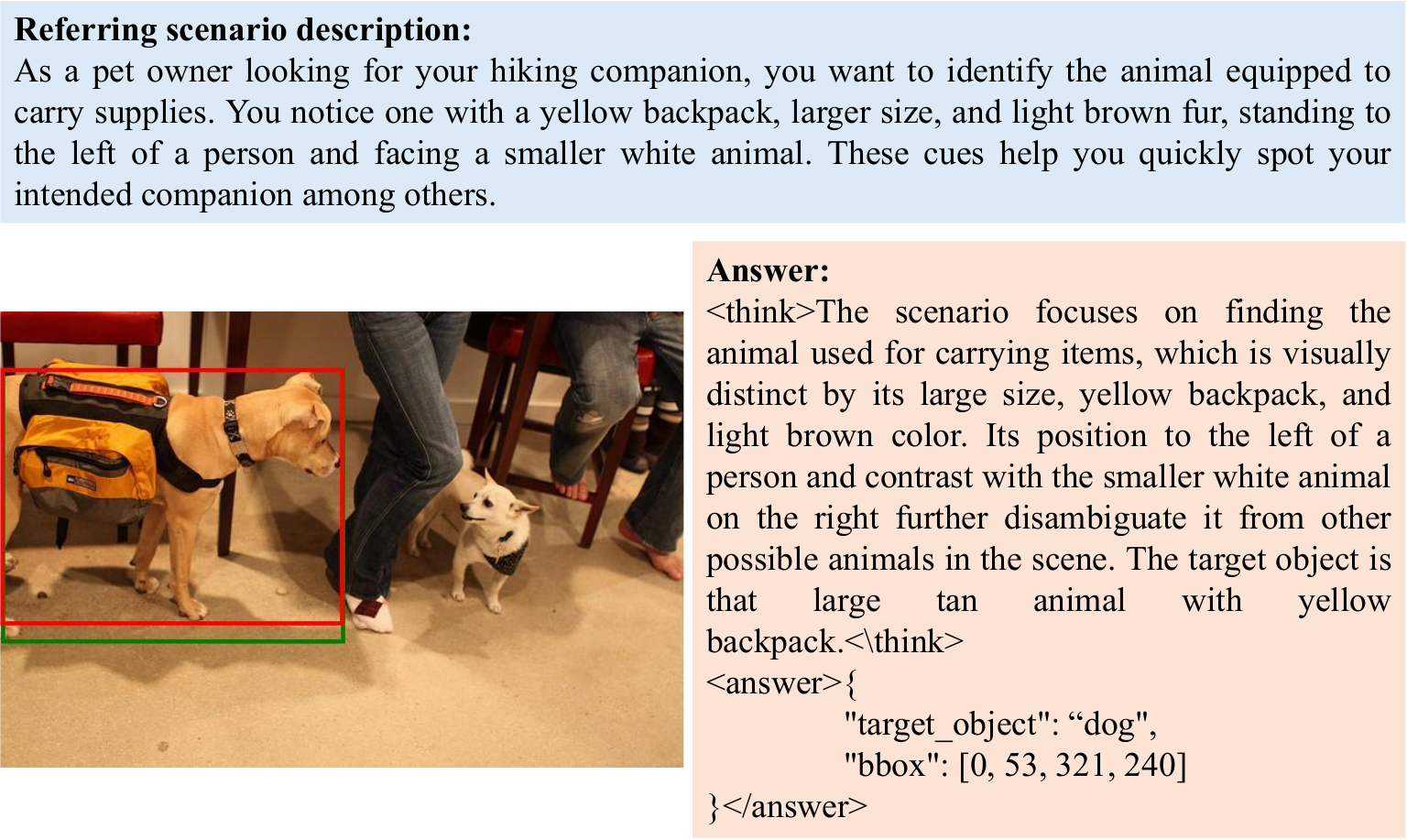}
    \includegraphics[width=0.48\linewidth]{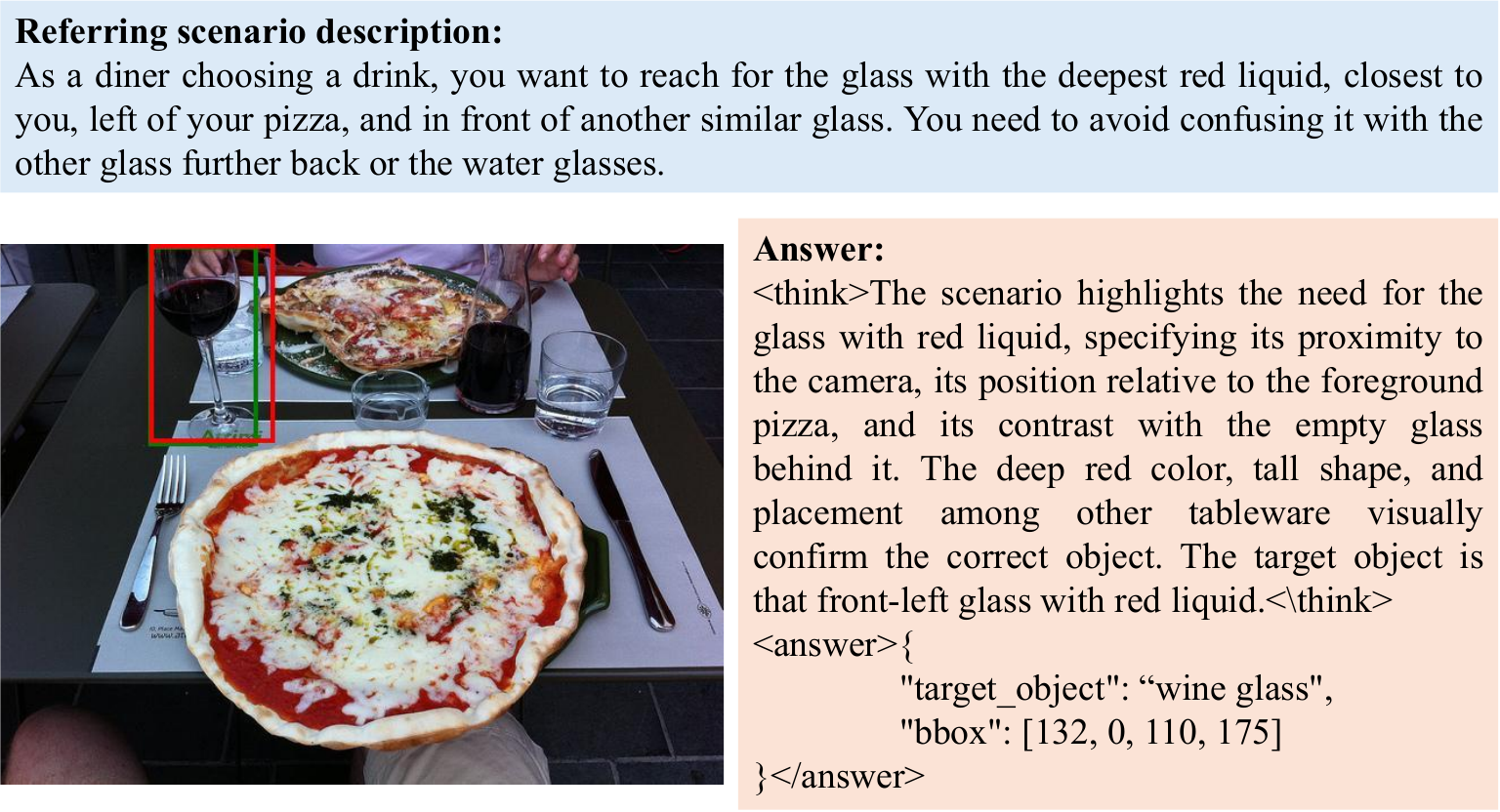}
    \caption{Qualitative examples of \method. Green for the GT box and red for the Predicted box.}
    \label{fig:qualitative_supp}
\end{figure*}

\figurename~\ref{fig:qualitative_supp} presents further qualitative examples illustrating how our \method–trained model uses scenario-aware reasoning to improve grounding performance. For each case, we show the input image together with the user scenario, the ground-truth target region, and the model’s prediction. The examples span challenging settings such as cluttered scenes, small or partially occluded objects, and pragmatically underspecified descriptions.

\section{Experimental Setup}
\label{app:exp_setup}

\paragraph{Base model.}
All experiments use Qwen2.5-VL-7B-Instruct~\cite{Qwen2.5-VL} 
as the backbone, with images resized to $512{\times}512$.

\paragraph{TP-SFT hyperparameters.}
AdamW optimizer, learning rate $5{\times}10^{-6}$, cosine 
schedule with warmup ratio $0.15$, 5 epochs, gradient 
accumulation steps $4$. The SFT split targets easy-purity 
$\delta_{\mathrm{easy}} \ge 0.70$.

\paragraph{IC-GRPO hyperparameters.}
We run two RL stages. Stage~1: AdamW, 
lr $1{\times}10^{-6}$, cosine schedule, warmup ratio 
$0.05$, 5 epochs, $K{=}6$ rollouts per prompt, generation 
batch 240, temperature $0.8$, top-$p$ $0.95$, max 
completion length 256. Stage~2: AdamW, lr $2{\times}10^{-6}$, 
cosine schedule, warmup ratio $0.05$, 2 epochs, $K{=}12$ 
rollouts per prompt, generation batch 288, temperature 
$0.9$, top-$p$ $0.92$, max completion length 160. Both 
stages use DeepSpeed ZeRO-3, gradient checkpointing, 
and bf16 precision.

\paragraph{Adaptive KL scheduler.}
Initial $\beta_0 = 2{\times}10^{-2}$. Stage~1: 
$\kappa_{\mathrm{tgt}}{=}0.13$, $\kappa_{\mathrm{tol}}{=}0.03$, 
$\mu_\uparrow{=}1.5$, $\mu_\downarrow{=}0.66$. Stage~2: 
$\kappa_{\mathrm{tgt}}{=}0.15$, $\kappa_{\mathrm{tol}}{=}0.03$, 
$\mu_\uparrow{=}1.6$, $\mu_\downarrow{=}0.66$. Both 
stages clip $\beta \in [5{\times}10^{-4},\,5{\times}10^{-2}]$.

\paragraph{Geometry reward parameters.}
IoU operating points $\tau_1{=}0.50$, $\tau_2{=}0.70$; 
logistic width $\kappa{=}0.03$; bonus magnitudes 
$\alpha_1{=}0.30$, $\alpha_2{=}0.50$; center coefficient 
$\alpha_c{=}0.02$ with bandwidth $\sigma_c{=}0.20$ 
(as a fraction of image diagonal); OOB penalty 
$\alpha_{\mathrm{oob}}{=}0.05$.

\paragraph{Category reward parameters.}
IoU gate threshold $\tau_g{=}0.30$, gate factor $g{=}0.5$; 
alias credit $\eta{=}0.80$; soft-overlap range 
$\rho_l{=}0.40$, $\rho_s{=}0.30$ (mapping Jaccard 
$\in[0,1]$ to $[0.40,\,0.70]$).

\paragraph{Format and structure reward parameters.}
$\gamma_{\mathrm{tag}}{=}0.25$, 
$\gamma_{\mathrm{key}}{=}0.75$, 
$\gamma_{\min}{=}{-}0.50$.

\paragraph{Reward weight schedule.}
Annealing completes by $p_{\mathrm{anneal}}{=}0.60$ of 
total training steps. Concrete start and late values are:

\begin{table}[h]
\centering
\small
\setlength{\tabcolsep}{6pt}
\caption{Reward weight schedules. Stage~2 weights are 
set by the \texttt{RewardWeightScheduler} from step 0, 
overriding the initialisation values.}
\label{tab:reward_weights}
\begin{tabular}{lcccc}
\toprule
& $w_{\mathrm{iou}}$ & $w_{\mathrm{cat}}$ 
& $w_{\mathrm{fmt}}$ & $w_{\mathrm{struct}}$ \\
\midrule
Stage~1 & 0.75 & 0.15 & 0.07 & 0.03 \\
Stage~2 (start) & 0.55 & 0.25 & 0.12 & 0.08 \\
Stage~2 (late)  & 0.75 & 0.20 & 0.04 & 0.01 \\
\bottomrule
\end{tabular}
\end{table}

\paragraph{Curriculum mixtures.}
Stage~1 RL uses 
$\Pi^{(1)}{=}(0.70,\,0.30,\,0.00)$; 
Stage~2 RL uses 
$\Pi^{(2)}{=}(0.20,\,0.60,\,0.20)$ 
for (easy, medium, hard) respectively.

\paragraph{Prompt-template ensemble.}
$T{=}8$ paraphrase templates (PTE-8).

\section{Limitations and Future Work}
\label{app:limitations}

RSC scenarios are LLM-generated rather than collected from real users. Although our ecological validity study shows that model rankings are preserved across human- and LLM-authored queries for the same instances, the synthetic scenarios exhibit less stylistic diversity than naturalistic language. 
RSC is also limited to single-target grounding on MS-COCO and LVIS images, which may not generalize to specialized domains or multi-object, video-based, and interactive settings. 
Future work could complement RSC with human-authored queries to validate ecological validity, extend the benchmark to diverse image 
domains, and explore multi-object and dialogue-grounded scenario grounding. 

\section{Broader Impact}
\label{app:broader_impact}

RSC is intended to support the development of multimodal models that reason over rich natural language, with applications in embodied assistants and accessibility tools. The benchmark inherits demographic and geographic biases from MS-COCO and LVIS, and LLM-generated scenarios may carry residual hallucinations or culturally specific assumptions despite quality controls. We release annotation logs and metadata to support bias analysis. RSC contains no personally identifiable information, and we do not anticipate direct misuse risks, though we encourage downstream users to consider the implications of scenario-based grounding capabilities in sensitive deployment contexts.

\end{document}